\documentclass{article} 
\usepackage{iclr2024_conference,times}

\usepackage{amsmath,amsfonts,bm}

\def\eqref#1{equation~\ref{#1}}

\def\1{\bm{1}}

\def\mA{{\bm{A}}}
\def\mB{{\bm{B}}}
\def\mC{{\bm{C}}}

\def\mX{{\bm{X}}}

\DeclareMathAlphabet{\mathsfit}{\encodingdefault}{\sfdefault}{m}{sl}
\SetMathAlphabet{\mathsfit}{bold}{\encodingdefault}{\sfdefault}{bx}{n}

\usepackage{hyperref}
\usepackage{url}
\usepackage{chemformula}

\usepackage{graphicx}%
\usepackage{multirow}%
\usepackage{amsmath,amssymb,amsfonts}%
\usepackage{amsthm}%
\usepackage{mathrsfs}%
\usepackage[title]{appendix}%
\usepackage{xcolor}%
\usepackage{textcomp}%
\usepackage{manyfoot}%
\usepackage{booktabs}%
\usepackage{listings}%

\usepackage{subfig}
\usepackage[nolist]{acronym}
\usepackage{microtype}      
\usepackage[linesnumbered,ruled]{algorithm2e}

\title{Würstchen: \\ An Efficient Architecture for Large-Scale Text-to-Image Diffusion Models}
\author{Pablo Pertinas\thanks{equal contribution}\\
Indpendent researcher, Sant Joan d’Alacant, Spain
\AND 
Dominic Rampas\textsuperscript{$\ast$}\\
Technische Hochschule Ingolstadt, Ingolstadt, Germany \\
Wand Technologies Inc., New York, USA \\
\AND 
Mats L. Richter\textsuperscript{$\ast$} \\
Université de Montréal, Montreal, Canada \\
Mila, Quebec AI Institute, Montreal, Canada \\
\AND 
Christopher J. Pal \\
Polytechnique Montréal, Montreal, Canada \\
Mila, Quebec AI Institute, Quebec, Canada \\
Canada CIFAR AI Chair
\AND
Marc Aubreville \\
Technische Hochschule Ingolstadt, Ingolstadt, Germany \\
}

\iclrfinalcopy 

\newcommand{\etal}{\textit{et al}.~}

\begin{document}

\maketitle

\begin{abstract}
    We introduce Würstchen, a novel architecture for text-to-image synthesis that combines competitive performance with unprecedented cost-effectiveness for large-scale text-to-image diffusion models.
A key contribution of our work is to develop a latent diffusion technique in which we learn a detailed but extremely compact semantic image representation used to guide the diffusion process. This highly compressed representation of an image provides much more detailed guidance compared to latent representations of language and this significantly reduces the computational requirements to achieve state-of-the-art results. Our approach also improves the quality of text-conditioned image generation based on our user preference study.
The training requirements of our approach consists of 24,602 A100-GPU hours -- compared to Stable Diffusion 2.1's 200,000 GPU hours.  
Our approach also requires less training data to achieve these results. Furthermore, our compact latent representations allows us to perform inference over twice as fast, slashing the usual costs and carbon footprint of a state-of-the-art (SOTA) diffusion model significantly, without compromising the end performance. In a broader comparison against SOTA models our approach is substantially more efficient and compares favorably in terms of image quality.
    We believe that this work motivates more emphasis on the prioritization of both performance and computational accessibility.
\end{abstract}

\begin{figure}[h]
\vspace{-0.5cm}

\centering\includegraphics[width=0.90\textwidth]{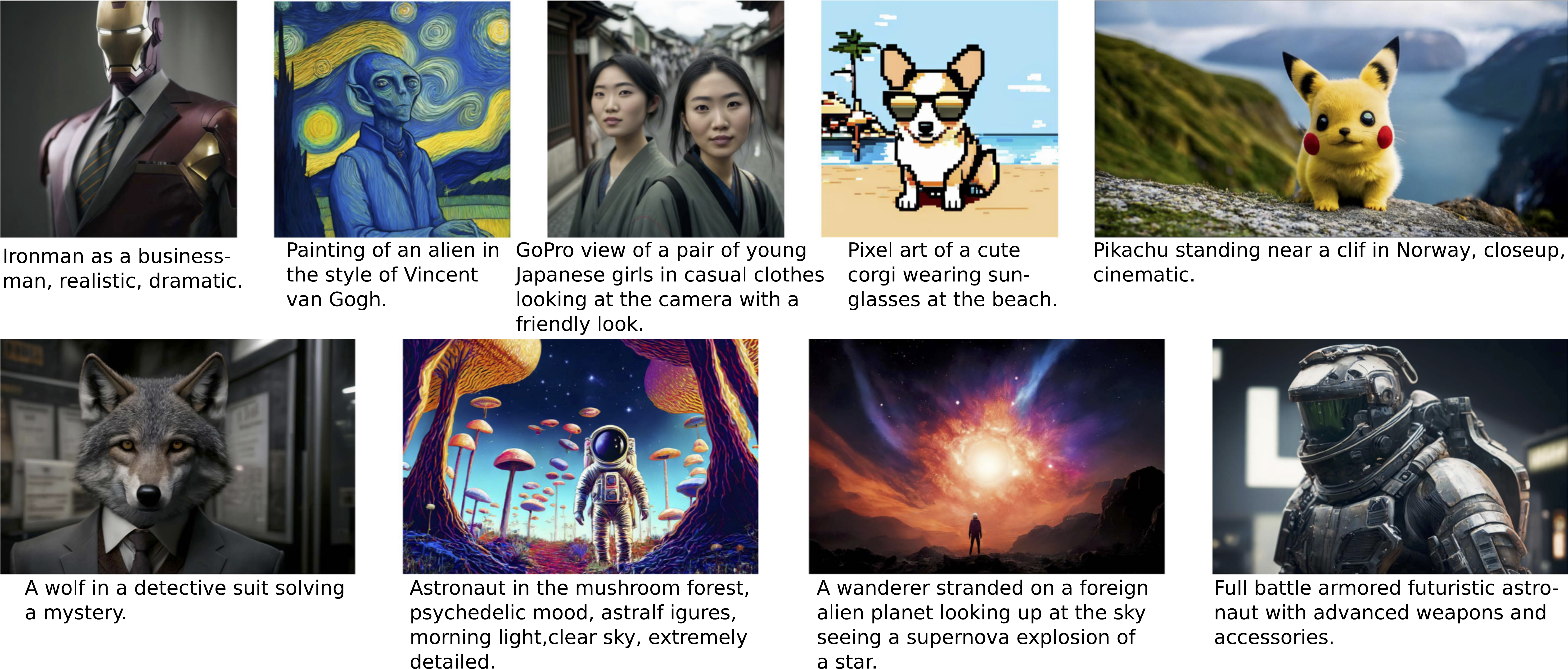}

\caption{Text-conditional generations using Würstchen. Note the various art styles and aspect ratios.}
\label{fig:main}
\vspace{-0.5cm}

\end{figure}


\section{Introduction}
\label{sec:intro}
State-of-the-art diffusion models \citep{ho2020denoising,saharia2022photorealistic,dalle2_ramesh2022hierarchical} have advanced the field of image synthesis considerably, achieving remarkable results that closely approximate photorealism. However, these foundation models, while impressive in their capabilities, carry a significant drawback: they are computationally demanding. For instance, \ac{SD} 1.4, one of the most notable models in the field, used 150,000 GPU hours for training \citep{sd14-model-card}. 
While more economical text-to-image models do exist \citep{ding2021cogview, ding2022cogview2, galip, tao2022df}, the image quality of these models can be considered inferior in terms of lower resolution and overall aesthetic features.


%
The core dilemma for this discrepancy is that increasing the resolution also increases visual complexity and computational cost, making image synthesis more expensive and data-intensive to train.
Encoder-based \acp{LDM} partially address this by operating on a compressed latent space instead of directly on the pixel-space \citep{rombach2022high}, but are ultimately limited by how much the encoder-decoder model can compress the image without degradation \citep{sizematters}.

\begin{figure}[bt]
\centering\includegraphics[width=0.6\textwidth]{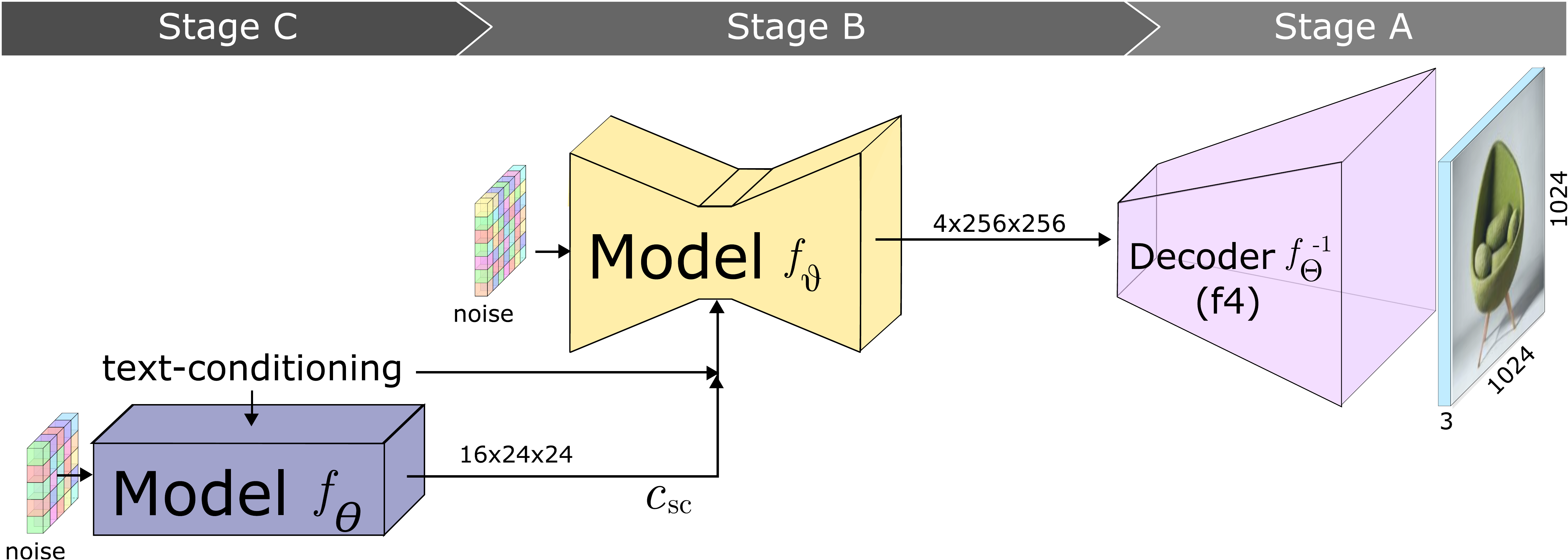}
\caption{Inference architecture for text-conditional image generation.}
\label{fig:inference}
\vspace{-0.5cm}
\end{figure}

Against this backdrop, we propose a novel three-stage architecture named "Würstchen", which drastically reduces the computational demands while maintaining competitive performance. 
We achieve this by training a diffusion model on a very low dimensional latent space with a high 
compression ratio of 42:1.
This very low dimensional latent-space is used to condition the second generative latent model, effectively helping it to navigate a higher dimensional latent space of a \ac{VQGAN}, which operates at a compression ratio of 4:1.
More concretely, the approach  uses three distinct stages for image synthesis (see Figure~\ref{fig:inference}): initially, a text-conditional \ac{LDM} is used to create a low dimensional latent representation of the image (Stage C). This latent representation is used to condition another \ac{LDM} (Stage B), producing a latent image in a latent space of higher dimensionality. Finally, the latent image is decoded by a \ac{VQGAN}-decoder to yield the full-resolution output image (Stage A).

Training is performed in reverse order to the inference (Figure~\ref{fig:training}): The initial training is carried out on Stage A and employs a \ac{VQGAN} to create a latent space. This compact representation facilitates learning and inference speed \citep{rombach2022high,chang2023muse,rampas2023novel}.
The next phase (Stage B) involves a first latent diffusion process \citep{rombach2022high}, conditioned on the outputs of a Semantic Compressor (an encoder operating at a very high spatial compression rate) and on text embeddings. This diffusion process is tasked to reconstruct the latent space established by the training of Stage A, which is strongly guided by the detailed semantic information provided by the Semantic Compressor.
Finally, for the construction of Stage C, the strongly compressed latents of the Semantic Compressor from Stage B are used to project images into the condensed latent space where a text-conditional \ac{LDM} \citep{rombach2022high} is trained.
The significant reduction in space dimensions in Stage C allows for more efficient training and inference of the diffusion model, considerably reducing both the computational resources required and the time taken for the process. 
 
Our proposed Würstchen model thus introduces a thoughtfully designed approach to address the high computational burden of current state-of-the-art models, providing a significant leap forward in text-to-image synthesis. With this approach we are able to train a 1B parameter Stage C text-conditional diffusion model within approximately 24,602 GPU hours, resembling a 8x reduction in computation compared to the amount \ac{SD} 2.1 used for training (200,000 GPU hours), while showing similar fidelity both visually and numerically. Throughout this paper, we provide a comprehensive evaluation of Würstchen's efficacy, demonstrating its potential to democratize the deployment \& training of high-quality image synthesis models.

Our main contributions are the following:
\begin{enumerate}
\item We propose a novel three-stage architecture for text-to-image synthesis at strong compression ratio, consisting of two conditional latent diffusion stages and a latent image decoder.
\item We show that by using a text-conditional diffusion model in a strongly compressed latent space we can achieve state-of-the-art model performance at a significantly reduced training cost and inference speed.
\item We provide comprehensive experimental validation of the model's efficacy based on automated metrics and human feedback. 
\item We are publicly releasing the source code and the entire suite of model weights.
\end{enumerate}

\section{Related Work}
\label{sec:rel_work}
\subsection{Conditional Image Generation}
The field of image generation guided by text prompts has undergone significant progression in recent years. Initial approaches predominantly leveraged \acp{GAN}~\citep{reed2016generative,zhang2017stackgan}. More recently, however, a paradigm shift in the field of image generation towards diffusion models~\citep{https://doi.org/10.48550/arxiv.1503.03585,ho2020denoising} has occurred. These approaches, in some cases, have not only met but even exceeded the performance of \acp{GAN} in both conditional and unconditional image generation \citep{dhariwal2021diffusion}. Diffusion models put forth a score-based scheme that gradually eliminates perturbations (e.g., noise) from a target image, with the training objective framed as a reweighted variational lower-bound. Next to diffusion models, another dominant choice for training text-to-image models is transformers.
In their early stages, transformer-based models utilized an autoregressive approach, leading to a significant slowdown in inference due to the requirement for each token to be sampled individually. Current strategies, however, employ a bidirectional transformer \citep{ding2022cogview2,maskgit_chang2022maskgit,chang2023muse} to address the challenges that traditional autoregressive models present. As a result, image generation can be executed using fewer steps, while also benefiting from a global context during the generative phase. Other recent work has shown that convolution-based approaches for image generation can yield similar results \citep{rampas2023novel}.

\subsection{Compressed Latent Spaces}
The majority of approaches in the visual modality of generative models use some way to train at a smaller space, followed by upscaling to high resolutions, as training at large pixel resolutions can become exponentially more expensive with the size of images. For text-conditional image generation, there are two established categories of approaches: encoder-based and upsampler-based.
\acp{LDM} \citep{rombach2022high}, DALL-E \citep{ramesh2021zero}, CogView \citep{ding2021cogview,ding2022cogview2}, MUSE \citep{chang2023muse} belong to the first category and employ a two-stage training process. Initially, an autoencoder \citep{rumelhart1985learning} is trained to provide a lower-dimensional, yet perceptually equivalent, representation of the data. This representation forms the basis for the subsequent training of a diffusion or transformer model. Eventually, generated latent representations can be decoded with the decoder branch of the autoencoder to the pixel space. The result is a significant reduction in computational complexity for the diffusion/sampling process and efficient image decoding from the latent space using a single network pass.
On the contrary, upsampler-based methods generate images at low resolution in the pixel space and use subsequent models for upscaling the images to higher resolution. UnClip \citep{dalle2_ramesh2022hierarchical} and Imagen \citep{saharia2022photorealistic} both generate images at 64x64 and upscale using two models to 256 and 1024 pixels. The former model is the largest in terms of parameter count, while the latter models are smaller due to working at higher resolution and only being responsible for upscaling.

\subsection{Conditional Guidance}
The conditional guidance of models in text-based scenarios is typically facilitated through the encoding of textual prompts via a pretrained language model. Two major categories of text encoders are employed: contrastive text encoders and uni-modal text encoders.
\ac{CLIP} \citep{clip_radford2021learning} is a representative of the contrastive multimodal models that strives to align text descriptions and images bearing semantic resemblance within a common latent space. A host of image generation methodologies have adopted a frozen \ac{CLIP} model as their exclusive conditioning method in recent literature. The hierarchical DALL-E 2 by \citet{dalle2_ramesh2022hierarchical} specifically harnesses \ac{CLIP} image embeddings as input for their diffusion model, while a 'prior' performs the conversion of \ac{CLIP} text embeddings to image embeddings. \ac{SD} \citep{rombach2022high}, on the other hand, makes use of un-pooled \ac{CLIP} text embeddings to condition its \ac{LDM}.
In contrast, the works of \citet{saharia2022photorealistic}, \citet{liu2022character} and \citet{chang2023muse} leverage a large, uni-modal language model such as T5~\citep{raffel2020exploring} or ByT5~\citep{xue2022byt5} that can encode textual prompts with notable accuracy, leading to image generations of superior precision in terms of composition, style, and layout.

\section{Method}
\begin{figure}[ht!]
\centering\includegraphics[width=0.6\textwidth]{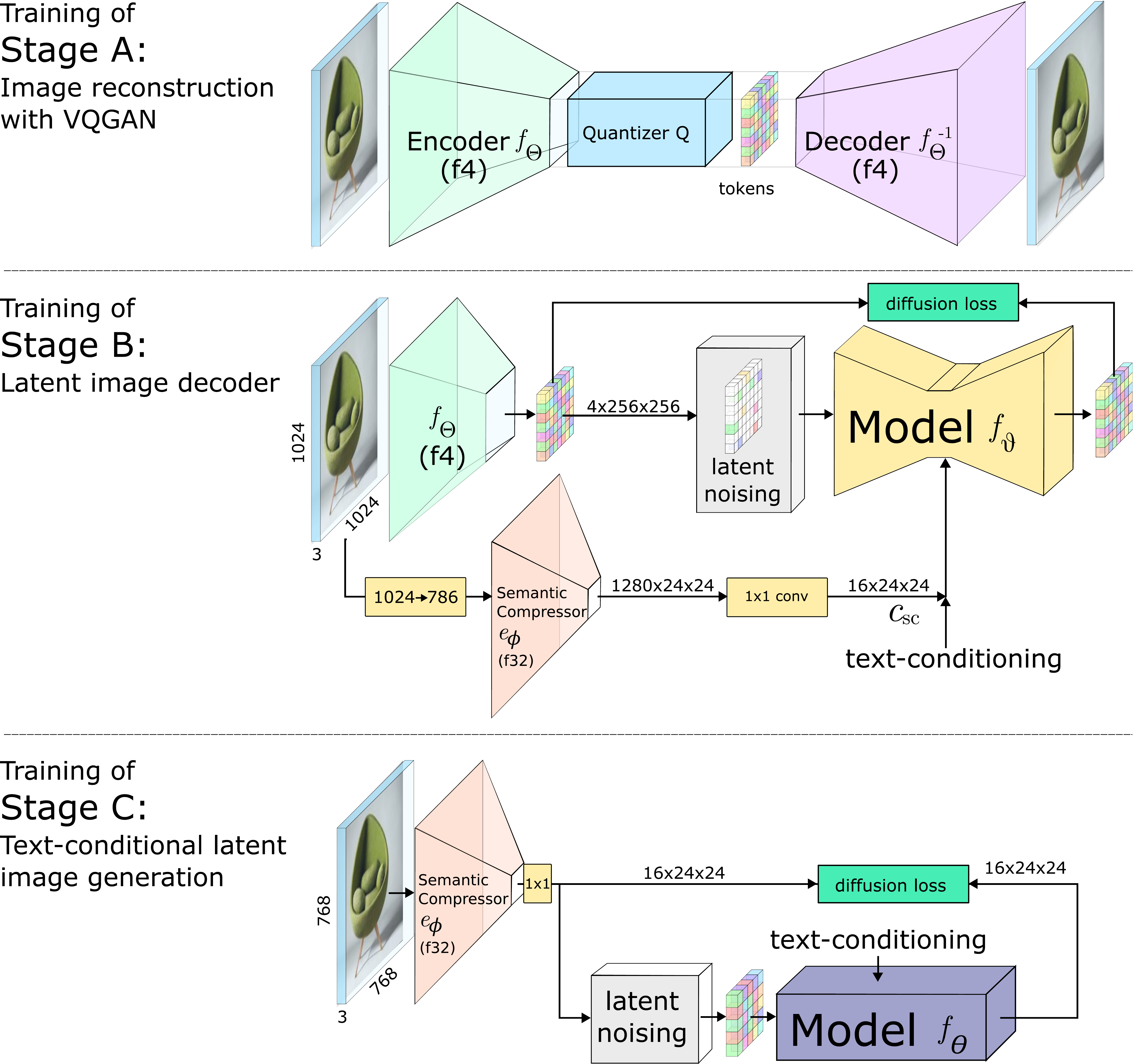}
\caption{Training objectives of our model. Initially, a VQGAN is trained. Secondly, Stage B is trained as a diffusion model inside Stage A's latent space. Stage B is conditioned on text-embeddings and the output of the Semantic Compressor, which produces strongly downsampled latent representations of the same image. Finally, Stage C is trained on the latents of the Semantic Compressor as a text-conditional \ac{LDM}, effectively operating on a compression ratio of $42:1$.}
\label{fig:training}
\end{figure}

Our method comprises three stages, all implemented as deep neural networks. For image generation, we first generate a latent image at a strong compression ratio using a text-conditional \ac{LDM} (Stage C). Subsequently, this representation is transformed to a less-compressed latent space by the means of a secondary model which is tasked for this reconstruction (Stage B). Finally, the tokens that comprise the latent image in this intermediate resolution are decoded to yield the output image (Stage A). The training of this architecture is performed in reverse order, starting with Stage A, then following up with Stage B and finally Stage C (see Figure~\ref{fig:training}). Text conditioning is applied on Stage C using CLIP-H \citep{ilharco_gabriel_2021_5143773}. Details on the training procedure can be found in Appendix \ref{sect:model_details}.

\label{sec:method}
\subsection{Stage A and B}
It is a known and well-studied technique to reduce the computational burden by compressing data into a smaller representation\citep{sizematters, Richter2021ShouldYG, maskgit_chang2022maskgit}. Our approach follows this paradigm, too, and makes use of Stages A \& B to achieve a notably higher compression than usual. Let $H\times W\times C$ be the dimensions of images. A spatial compression maps images to a latent representation with a resolution of 
$h\times w\times z$ with $h = H/f, w = W/f$, where $f$ defines the compression rate. Common approaches for modeling image synthesis use a one-stage compression between $f$4 and $f$16 \citep{vqgan_esser2021taming,chang2023muse,rombach2022high}, with higher factors usually resulting in worse reconstructions. Our Stage A consists of a $f$4 VQGAN \citep{vqgan_esser2021taming} with parameters $\Theta$ and initially encodes images $\mX \in \mathbb{R}^{3\times1024\times 1024}$ into $256\times256$ discrete tokens from a learned codebook of size 8,192.
\[ \mX_{q} = f_{\Theta}(\mX) \]
The network is trained as described by Esser \etal and tries to reconstruct the image based on the quantized latents, so that:
\[ f_\Theta^{-1}\left(f_\Theta\left( \mX\right)\right) = f_\Theta^{-1}  \left(\mX_q\right)  \approx \mX\]
where $f_\Theta^{-1}$ resembles the decoder part of the VQGAN. 

Afterward, the quantization is dropped from Stage A, and Stage B is trained in the unquantized latent space of the Stage A-encoder as a conditioned \ac{LDM}. 
In stage B, we utilize a Semantic Compressor, i.e., an encoder-type network that is tasked to create latent representations at a strong spatial compression rate that can be used to create a latent representation to guide the diffusion process. 
 The unquantized image embeddings are noised following an LDM training procedure. The noised representation $\tilde{\mX}_{t}$, together with the visual embeddings from the Semantic Compressor, $\mC_\mathrm{sc}$, text conditioning $\mC_\mathrm{text}$ and the timestep $t$ are given to the model.

The highly compressed visual embeddings extracted by the Semantic Compressor will act as an interface for Stage C, which will be trained to generate them. The embeddings will have a shape of $ \mathbb{R}^{1280\times24\times24}$ obtained by encoding images with shape $\mX \in \mathbb{R}^{3\times786\times 786}$. 
We use simple bicubic interpolation for the resizing of the images from $1024\times 1024$ to $786 \times 786$, which is a sufficiently high resolution to fully utilize the parameters of the Semantic Compressor \citep{Richter2023, richter2022receptive}, while also reducing the latent representation size.
Moreover, we further compress the latents with a $1\times 1$ convolution that normalizes and projects the embeddings to $\mC_\mathrm{sc} \in \mathbb{R}^{16\times24\times24}$. This compressed representation of the images is given to the Stage~B decoder as conditioning to guide the decoding process.

\[ \bar{\mX}_{0} = f_{\vartheta}(\tilde{\mX}_{t}, \mC_\mathrm{sc}, \mC_\mathrm{text}, t) \]

By conditioning Stage B on low-dimensional latent representations, we can effectively decode images from a $\mathbb{R}^{16\times24\times24}$ latent space to a resolution of $\mX \in \mathbb{R}^{3\times1024\times 1024}$, resulting in a total spatial compression of \textbf{42:1}. 

We initialized the Semantic Compressor with weights pre-trained on ImageNet, which, however, does not capture the broad distribution of images present in large text-image datasets and is not well-suited for semantic image projection, since it was trained with an objective to discriminate the ImageNet categories. Hence we updated the weights of the Semantic Compressor during training, establishing a latent space with high-precision semantic information. We use Cross-Attention \citep{transformers_vaswani2017attention} for conditioning and project $\mC_\mathrm{sc}$ (flattened) to the same dimension in each block of the model and concatenate them. Furthermore, during training Stage B, we intermittently add noise to the Semantic Compressor's embeddings, to teach the model to understand non-perfect embeddings, which is likely to be the case when generating these embeddings with Stage C. Lastly, we also randomly drop $\mC_\mathrm{sc}$ to be able to sample with classifier-free-guidance \citep{classifier_free_guidance_ho2022classifier} during sampling.

\subsection{Stage C}
After Stage A and Stage B were trained, training of the text-conditional last stage started. In our implementation, Stage C consists of 16 ConvNeXt-block \citep{liu2022convnet} without downsampling, text and time step conditionings are applied after each block via cross-attention. We follow a standard diffusion process, applied in the latent space of the finetuned Semantic Compressor. Images are encoded into their latent representation 
$\mX_\mathrm{sc} = \mC_\mathrm{sc}$, representing the target. The latents are noised by using the following forward diffusion formula:
\[ \mX_{\mathrm{sc}, t} = \sqrt{\bar\alpha_{t}} \cdot \mX_\mathrm{sc} + \sqrt{1 - \bar\alpha_{t}} \cdot \epsilon \]
where $\epsilon$ represents noise from a zero mean unit variance normal distribution. 
We use a cosine schedule \citep{nichol2021improved} to generate $\bar\alpha_t$ and use continuous timesteps.
The diffusion model takes in the noised embeddings $\mX_{\mathrm{sc}, t}$, the text conditioning $\mC_\mathrm{text}$ and the timestep $t$. The model returns the prediction for the noise in the following form:
$$\bar\epsilon = \frac{\mX_{\mathrm{sc},t} - \mA}{\mid 1-\mB \mid + 1e^{-5}}$$
with
\[ \mA, \mB = f_{\theta}(\mX_{\mathrm{sc},t}, \mC_\mathrm{text}, t) \]

We decided to formulate the objective as such, since it made the training more stable. We hypothesize this occurs because the model parameters are initialized to predict $\mathbf{0}$ at the beginning, enlarging the difference to timesteps with a lot of noise. By reformulating to the $\mA$ \& $\mB$ objective, the model initially returns the input, making the loss small for very noised inputs. We use the standard mean-squared-error loss between the predicted noise and the ground truth noise. Additionally, we employ the p2 loss weighting \citep{choi2022perception}:
$$p_2(t) \cdot \mid\mid \epsilon - \bar\epsilon \mid\mid^2$$
where $p_2(t)$ is defined as $\frac{1 - \bar\alpha_{t}}{1 + \bar\alpha_{t}}$, making higher noise levels contribute more to the loss. Text conditioning $\mC_\mathrm{text}$ are dropped randomly for 5\% of the time and replaced with a null-label in order to use classifier-free-guidance \citep{classifier_free_guidance_ho2022classifier}

\subsection{Image Generation (Sampling)}
A depiction of the sampling pipeline can be seen in Figure~\ref{fig:inference}.
Sampling starts at Stage C, which is primarily responsible for image-synthesis (see Appendix \ref{sect:workload}), from initial random noise $\mX_{\mathrm{sc}, \tau_{C}} = \mathcal{N}(0, \mathbf{I})$. We use the DDPM \citep{ho2020denoising} algorithm to sample the Semantic Compressor latents conditioned on text-embeddings. To do so, we run the following operation for $\tau_{C}$ steps:

\[ \hat{\mX}_{\mathrm{sc}, t-1} = \frac{1}{\sqrt{\alpha_{t}}} \cdot (\hat{\mX}_{\mathrm{sc}, t} - \frac{1-\alpha_{t}}{\sqrt{1-\bar\alpha_{t}}}\bar\epsilon) + \sqrt{(1-\alpha_{t}) \frac{1-\bar\alpha_{t-1}}{1-\bar\alpha_{t}}}\epsilon \]

We denote the outcome as $\bar{\mX}_\mathrm{sc}$ which is of shape $16\times24\times24$. This output is flattened to a shape of $576\times16$ and given as conditioning, along with the same text embeddings used to sample $\bar{\mX}_\mathrm{sc}$, to Stage B. This stage operates at $4\times256\times256$ unquantized VQGAN latent space. We initialize $\mX_{q, \tau_B}$ to random tokens drawn from the VQGAN codebook. We sample $\tilde{\mX}$ for $\tau_B$ steps using the standard \ac{LDM} scheme. 

\[ \tilde \mX_{t-1} = f_{\vartheta}(\tilde \mX_{t}, \mC_\mathrm{sc}, \mC_\mathrm{text}, t) \]

Finally $\tilde  {\mX}$ is projected back to the pixel space using the decoder $f_\Theta^{-1}$ of the VQGAN (Stage A):

\[ \bar{\mX} = f_{\Theta}^{-1}(\tilde {\mX}) \]

\subsection{Model Decisions}
 Theoretically, any feature extractor could be used as backbone for the Semantic Compressor. 
 However, we hypothesize that it is beneficial to use a backbone that already has a good feature representation of a wide variety of images. Furthermore, having a small Semantic Compressor makes training of Stage B \& C faster. Finally, the feature dimension is vital. If it is excessively small, it may fail to capture sufficient image details or will underutilize parameters \citep{richter2022receptive}; conversely, if it is overly large, it may unnecessarily increase computational requirements and extend training duration \citep{sizematters}.
For this reason, we decided to use an ImageNet1k pre-trained EfficientV2 (S) as the backbone for our Semantic Compressor, as it combines high compression with well generalizing feature representations and computational efficiency.

Furthermore, we deviate in Stage C from the U-Net standard architecture. As the image is already compressed by a factor of 42, and we find further compression harmful to the model quality. Instead, the model is a simple sequence of 16 ConvNeXt blocks \citep{liu2022convnet} without downsampling. Time and text conditioning is applied after each block.


\section{Experiments and Evaluation}
\label{sec:experiments}
To demonstrate Würstchen's capabilities on text-to-image generation, we trained an 18M parameter Stage A, a 1B parameter Stage B and a 1B parameter Stage C. We employed an EfficientNet2-Small as Semantic Compressor \citep{tan2020efficientnet} during training. Stage B and C are conditioned on un-pooled CLIP-H \citep{ilharco_gabriel_2021_5143773} text-embeddings. The setup is designed to produce images of variable aspect ratio with up to $1538$ pixels per side. All stages were trained on subsets of the improved-aesthetic LAION-5B \citep{schuhmann2022laion} dataset. 

All the experiments use the standard DDPM \citep{ho2020denoising} algorithm to sample latents in Stage B and C. Both stages also make use of classifier-free-guidance \citep{classifier_free_guidance_ho2022classifier} with guidance scale $w$. We fix the hyperparameters for Stage B sampling to $\tau_B=12$ and $w=4$, Stage C uses $\tau_{C}=60$ for sampling.
Images are generated using a $1024 \times 1024$ resolution.

\paragraph{Baselines}
To better assess the efficacy of our architecture, we additionally train a U-Net-based 1B parameter \ac{LDM} on \ac{SD} 2.1 first stage and text-conditioning model.
We refer to this model as Baseline LDM, it is trained for $\approx$ 25,000 GPU-hours (same as Stage C) using an $512 \times 512$ input resolution.

Additionally, we evaluate our model against various state-of-the-art models that were publicly available at the time of writing (see Tables \ref{tab:pic-score} and Table \ref{tab:fid-comp}). All these models were used in their respective default configuration for text-to-image synthesis. Whenever possible, the evaluation metrics published by the original authors were used.

\paragraph{Evaluation Metrics}
We used the  Fréchet Inception Distance (FID) \citep{heusel2018gans} and Inception Score (IS) to evaluate all our models on COCO-30K, similar to \citep{galip, ding2021cogview, ding2022cogview2}. 
For evaluating the FID score, all images were downsampled to $256 \times 256$ pixels to allow for a fair comparison between other models in the literature.
However, both metrics suffer from inconsistencies and are known to be not necessarily well correlated with the aesthetic quality perceived by humans (\citet{podell2023sdxl, ding2021cogview, ding2022cogview2}, see also Appendix \ref{sec:fid}).
For this reason, we chose PickScore \citep{kirstain2023pickapic} as our primary automated metric. 
PickScore is designed to imitate human preferences, when selecting from a set of images given the same prompt.
We applied PickScore to compare Würstchen to various other models on various datasets.
We provide the percentage of images, where PickScore preferred the image of Würstchen over the image of the other model.
To also evaluate the environmental impact of our model we estimated the carbon emitted during training based on the work of \citep{patterson2021carbon}.

Finally, we also conducted a study with human participants, where the participants chose between two images from two different models given the prompt.

\paragraph{Datasets}
To assess the zero-shot text-to-image capabilities of our model, we use three distinct sets of captions alongside their corresponding images.
The COCO-validation is the de-facto standard dataset to evaluate the zero-shot performance for text-to-image models.
For MS COCO we generate 30,000 images based on prompts randomly chosen from the validation set. We refer to this set of images as COCO30K.
Since the prompts of MS COCO are quite short and frequently lack detail, we also generate 5,000 images from the Localized Narrative MS COCO subset, we refer to his dataset as Localized Narratives-COCO-5K.
Finally, we also use Parti-prompts \citep{yu2022scaling}, a highly diverse set of 1633 captions, which closely reflects the usage scenario we intend for our model.


\subsection{Automated Text-to-Image Evaluation}
\label{sec:eval}
\begin{figure}
\centering\includegraphics[width=.6\textwidth]{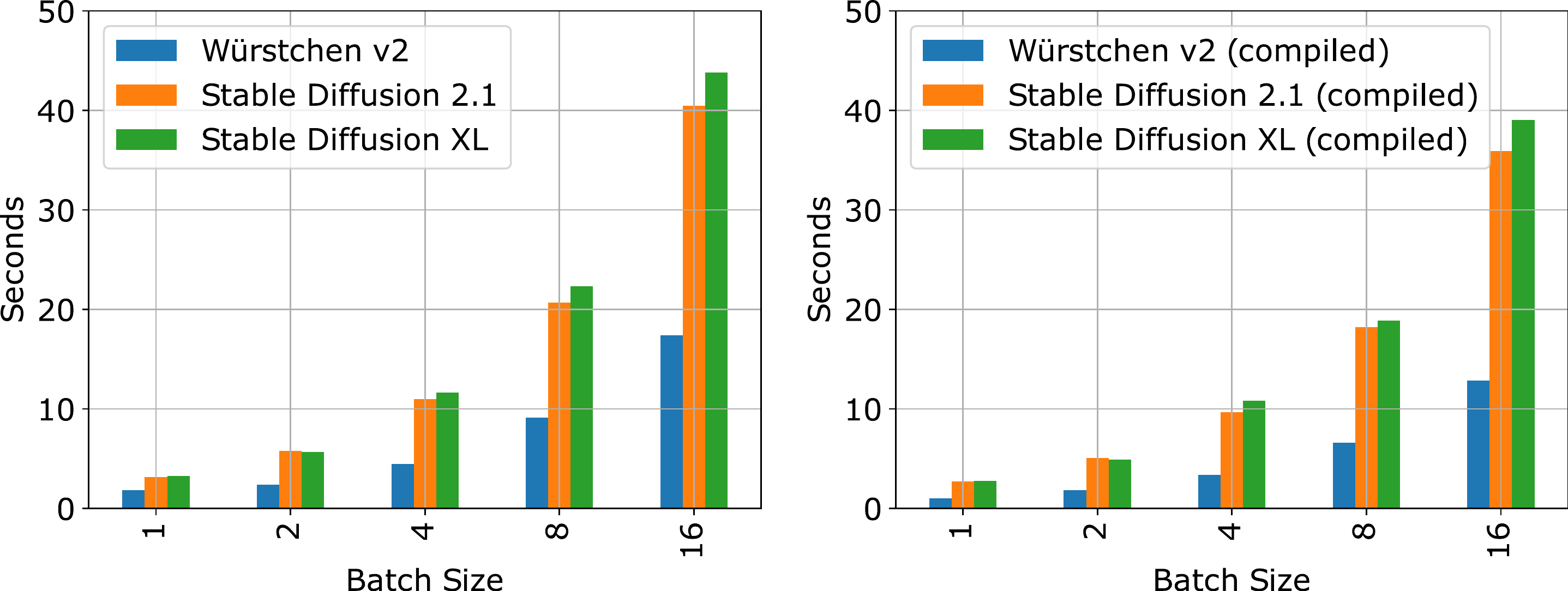}
\caption{Inference time for $1024\times 1024$ images on an A100-GPUs. Left plot shows performance without specific optimization, right plot shows performance using \texttt{torch.compile().}}
\label{fig:inference_speed}
\end{figure}

We evaluate the quality of the generated images using automated metrics in comparison to other, publicly available models (see Appendix \ref{sec:collages} for random examples).
The PickScores in Table \ref{tab:pic-score} paint a consistent picture over the three datasets the models were evaluated on. Würstchen is preferred very significantly over smaller models like DF-GAN and GALIP, which is expected. The \ac{LDM} is outperformed dramatically in all cases, highlighting that the architecture had a significant impact on the model's computational training efficiency. \textbf{Würstchen is also preferred in all three scenarios over \ac{SD} 1.4 and 2.1, despite their significantly higher compute-budget at a similar model-capacity.}
While \ac{SD} XL is still superior in image quality, our inference speed is significantly faster (see Figure \ref{fig:inference_speed}). This comparison is not entirely fair, as it's a higher capacity model and its data and compute budget is unknown. For this reason, we are omitting \ac{SD} XL from the following experiments.

While we achieve a higher Inception Score (IC) on COCO30K compared to all other models in our broader comparison in Table \ref{tab:fid-comp} also shows a relatively high FID on the same dataset. While still outperforming larger models like CogView2 \citep{ding2022cogview2} and our Baseline LDM, the FID is substantially lower compared to other state-of-the-art models.
We attribute this discrepancy to high-frequency features in the images.
During visual inspections we find that images generates by Würstchen tend smoother than in other text-to-image models. This difference is most noticeable in real-world images like COCO, on which we compute the FID-metric.

\begin{table}[t]

\caption{Evaluation of Image Quality on MS-COCO and Localized Narratives \citep{PontTuset_eccv2020} using the PickScore \citep{kirstain2023pickapic} to binary select images generated from the same captions by two different models. \textbf{Würstchen outperforms all models of equal and smaller size, despite Stable Diffusion models using a significantly higher compute budget.}}
\resizebox{\linewidth}{!}{
\begin{tabular}{|c|c|c|c|c|c|c|}
\hline
\multicolumn{1}{|c|}{} & \multicolumn{6}{c|}{PickScore(COCO-30k) $\uparrow$} \\
\hline
\textbf{Model} & Baseline LDM (ours)  & DF-GAN & GALIP & SD 1.4 & SD 2.1 & SD XL \\
\textbf{(train cost)}& ($\approx$25,000 gpu-h) & - & - & (150.000 gpu-h) & (200.000 gpu-h) & - \\ \hline

Würstchen & \multirow{2}{*}{\textbf{96.5\%}} & \multirow{2}{*}{\textbf{99.8\%}} & \multirow{2}{*}{\textbf{98.1\%}} & \multirow{2}{*}{\textbf{78.1\%}} & \multirow{2}{*}{\textbf{64.4\%}} & \multirow{2}{*}{39.4\%} \\ 
(24,602 gpu-h) &  &  &  &  & &\\ \hline
& \multicolumn{6}{c|}{PickScore (Localized Narratives-COCO-5K) $\uparrow$} \\ \hline
\textbf{Model} & Baseline LDM (ours) & DF-GAN & GALIP & SD 1.4 & SD 2.1 & SD XL \\
\textbf{(train cost)} & ($\approx$25,000 gpu-h) & - & - & (150.000 gpu-h) & (200.000 gpu-h) & - \\
\hline
Würstchen  & \multirow{2}{*}{\textbf{96.6\%}} & \multirow{2}{*}{\textbf{98.0\%}} & \multirow{2}{*}{\textbf{95.5\%}} & \multirow{2}{*}{\textbf{79.9\%}} & \multirow{2}{*}{\textbf{70.0\%}} & \multirow{2}{*}{39.1\%} \\ 
(24,602 gpu-h) &  &  &  &  & & \\ 
\hline
& \multicolumn{6}{c|}{PickScore (Parti-prompts) $\uparrow$} \\ \hline
\textbf{Model} & Baseline LDM (ours) & DF-GAN & GALIP & SD 1.4 & SD 2.1 & SD XL \\
\textbf{(train cost)} & ($\approx$25,000 gpu-h) & - & - & (150.000 gpu-h) & (200.000 gpu-h) & - \\
\hline
Würstchen  & \multirow{2}{*}{\textbf{98.6\%}} & \multirow{2}{*}{\textbf{99.6\%}} & \multirow{2}{*}{\textbf{97.9\%}} & \multirow{2}{*}{\textbf{82.1\%}} & \multirow{2}{*}{\textbf{74.6\%}} & \multirow{2}{*}{39.0\%} \\ 
(24,602 gpu-h) &  &  &  &  & & \\ 
\hline
\end{tabular}
}
\label{tab:pic-score}

\vspace{-0.7 cm}

\end{table}

\begin{table}[b]
\caption{Comparison to other architectures. $^*$ computed from own evaluation. $\dagger$ based on official model cards \citep{sd14-model-card, sd21-model-card}.
}
\resizebox{\linewidth}{!}{
\begin{tabular}{|l|r|r|r|r|c|r|r|r|}
\hline
Model            & Params & \parbox{1.5cm}{Sampling Steps} & \parbox{1.2cm}{FID $\downarrow$ $@256^2$} & \parbox{1.1cm}{IS $\uparrow$ $@299^2  $}   & \parbox{1.1cm}{Open Source} &\parbox{1.8cm}{GPU Hours @ A100 $\downarrow$}    & \parbox{1.4cm}{Train $\downarrow$ Samples}  & \parbox{2.5cm}{Est. Carbon Em. [kg \ch{CO2} eq.]}\\ 

\hline
GLIDE \citep{glide_nichol2021glide}            & 3.5B       & 250            & 12.24 &   & -- & --  &  -- & --\\ 
Make-A-Scene \citep{gafni2022make}     & 4B         & 1024           & 11.84 &       & -- & --  & -- & --\\ 
Parti \citep{https://doi.org/10.48550/arxiv.2206.10789}          & 20B        & 1024            & \textbf{7.23} & &  -- & --  & -- & --\\  
CogView \citep{ramesh2021zero}           & 4B        & 1024            & 27.1 & 22.4 &  \checkmark & --  & -- & --\\ 
CogView2 \citep{ding2022cogview2}           & 6B        & -            & 24.0 & 25.2 &  - & --  & -- & --\\ 
\hline

DF-GAN \citep{tao2022df}           & 19M        & -            & 19.3 & 18.6 &\checkmark & --  & -- & --\\ 
GALIP \citep{galip}           & 240M        & -            & 12.5 & 26.3* &\checkmark & --  & -- & --\\ 
DALL-E \citep{ramesh2021zero}           & 12B        & 256            & 17.89 & 17.9 & -- & --  & -- & --\\
LDM \citep{rombach2022high} & 1.45B & 250 & 12.63 & 30.3 &\checkmark & --  & -- & --\\

\hline
Baseline LDM (ours) & 0.99B & 60 & 43.5* & 20.1* & - & $\approx$25,000 & & $\approx$2,300 \\
Würstchen (ours) & 0.99B & 60 & 23.6* & \textbf{40.9*} & \checkmark & \textbf{24,602} & \textbf{1.42B} &  \textbf{2,276} \\
SD 1.4 \citep{rombach2022high} &    0.8B     &     50       & 16.2*  & 40.6*  &\checkmark &  150,000 $\dagger$  & 4.8B $\dagger$ & 11,250 $\dagger$ \\
SD 2.1 \citep{rombach2022high} &    0.8B     &     50       & 15.1* & 40.1* & \checkmark &  200,000 $\dagger$ & 3.9B $\dagger$ & 15,000 $\dagger$ \\
SD XL \citep{podell2023sdxl} & 2.6B & 50 & $>$ 18 & -- & \checkmark & -- & -- & --\\
\hline
\end{tabular}
}
\label{tab:fid-comp}
\end{table}

\subsection{Human Preference Evaluation}

While most metrics evaluated in the previous section are correlated with human preference \citep{kirstain2023pickapic,heusel2018gans,salimans2016improved}, we follow the example of other works and also conducted two brief studies on human preference. 
To simplify the evaluation, we solely compared Würstchen against \ac{SD} 2.1, its closest capacity and performance competitor, and evaluated the human preference between these two models following the setup of \cite{kirstain2023pickapic}.
In total, we conducted two studies using the generated images from Parti-prompts and COCO30K images. Participants were presented randomly chosen pairs of images in randomized order. For each pair the participants selected a preference for one or neither of the images (see Appendix \ref{sec:preference} for details). In total, 3343 (Parti-prompts) and 2262 (COCO Captions) comparisons by 90 participants were made.

\begin{figure}[t!]%
    \centering
    \subfloat[\centering Overall Preference]{{\includegraphics[width=0.25\textwidth]{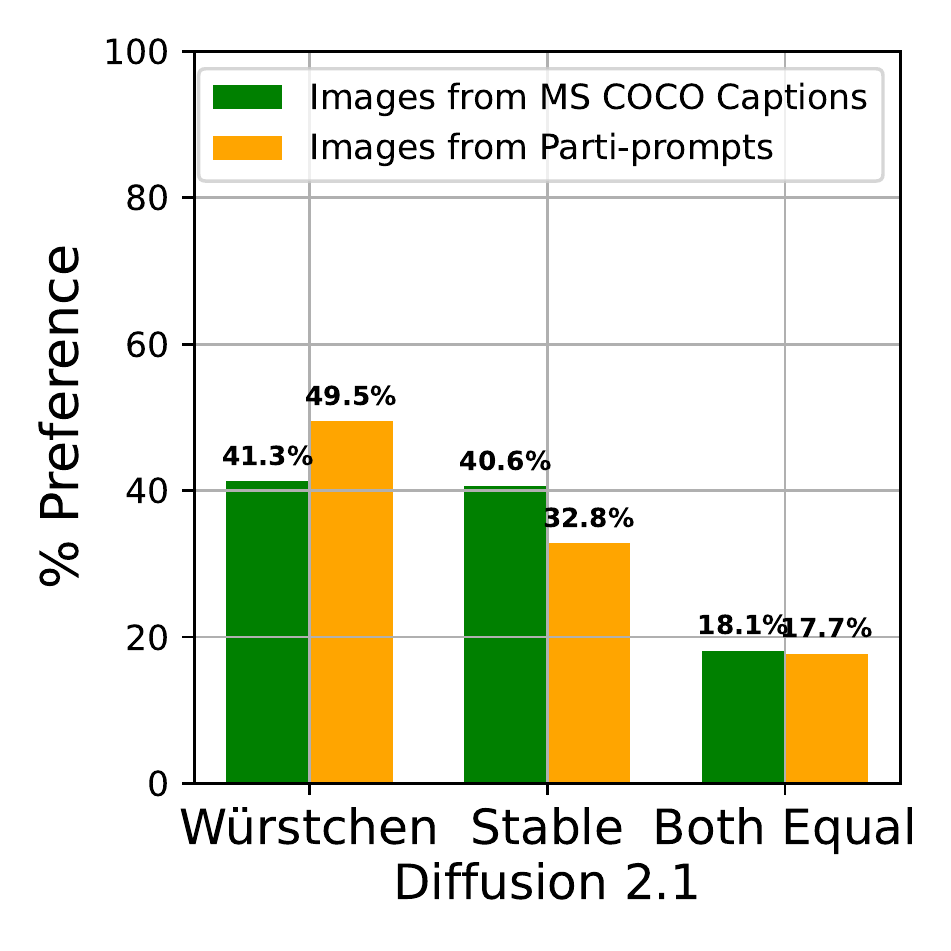} }}%
    \subfloat[\centering Individual Preference]{{\includegraphics[width=0.25\textwidth]{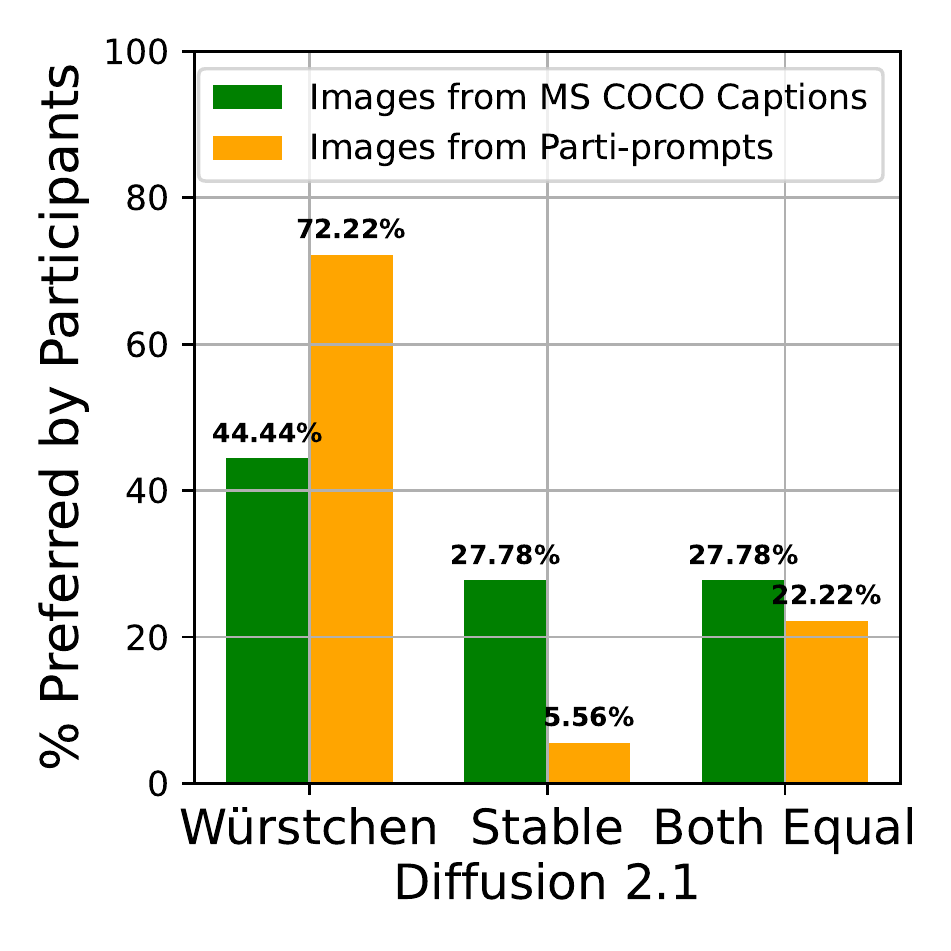} }}%
    \subfloat[\centering Histogram (MS COCO)]{{\includegraphics[width=0.25\textwidth]{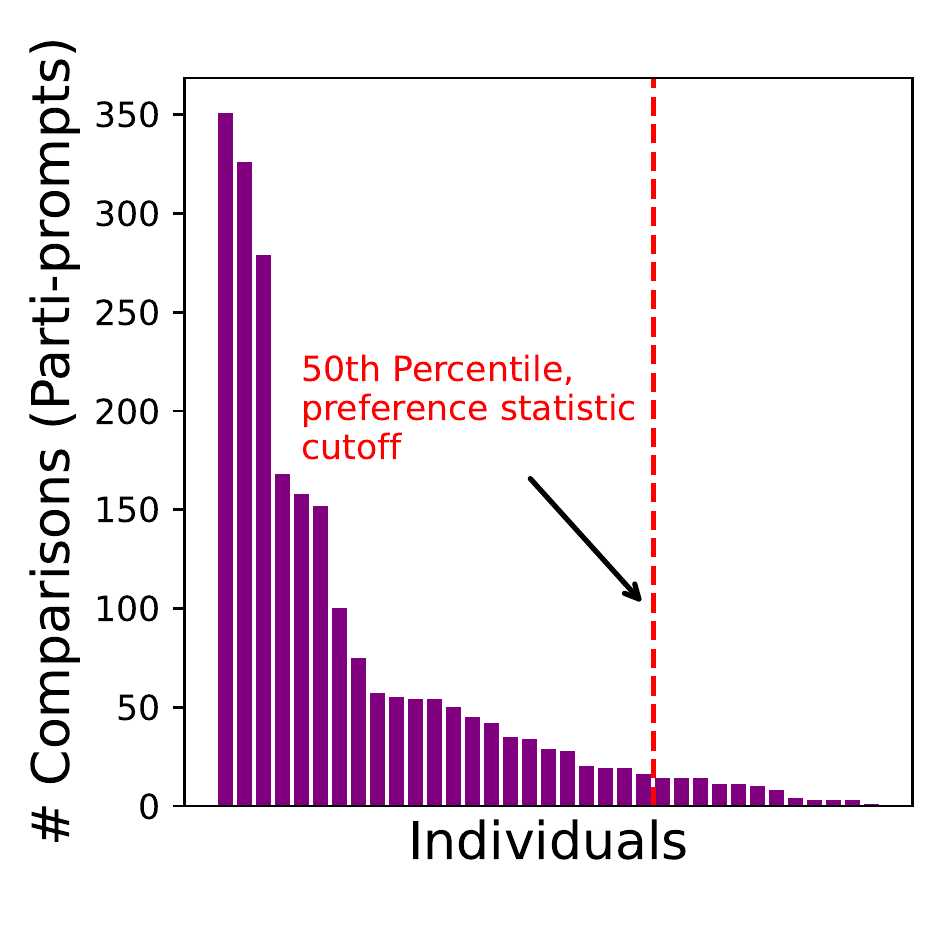} }}%
    \subfloat[\centering Histogram (Parti)]{{\includegraphics[width=0.25\textwidth]{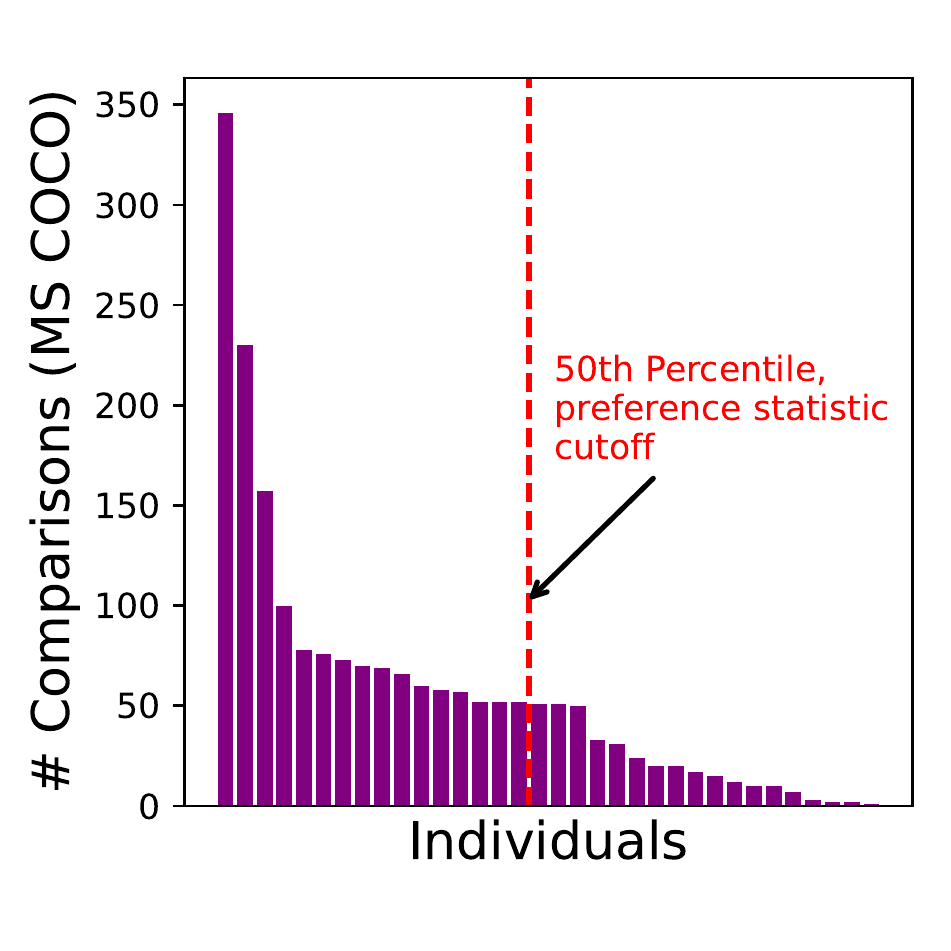} }}%
    \caption{Overall human preferences (left) and by users (middle). The preference by users considered only users with a large number of comparisons (right).}%
    \label{fig:study_results}%
    \vspace{-0.7cm}
    
\end{figure}

We evaluate results in two distinct ways. First, by counting the total number of preferences independent of user-identity. In Figure \ref{fig:study_results} (a) we can see that images generated by our model on Parti-prompts were clearly preferred. This is important to us, since Parti-prompt closely reflects the intended use case of the model. However, for MS-COCO this statistic is inconclusive.
We hypothesize that this is due to the vague prompts generating a more diverse set of images, making the preference more subject to personal taste, biasing this statistics towards users that completed more comparisons (Figure \ref{fig:study_results} (c, d)). 
For this reason, we conducted a second analysis, where we evaluated the personally preferred model for each individual. 
In an effort to only include participants that completed a representative number of comparisons, we only include users from the upper 50th percentile and above. By doing so, we include only individuals with at least 30 (MS-COCO) and 51 (Parti-prompts) comparisons in the statistic. 
Under these circumstances, we observed a light preference for MS-COCO in favor of Würstchen and a strong preference for our model on Parti-prompts (Figure \ref{fig:study} (b)).
In summary, the human preference experiments confirm the observation made in the PickScore experiments. While the real-world results were in-part less decisive, \textbf{the image generation quality of Würstchen was overall preferred by the participants of both studies over \ac{SD} 2.1}.

\subsection{Efficiency}
Table~\ref{tab:fid-comp} shows the computational costs for training Würstchen compared to the original \ac{SD} 1.4 and 2.1. Based on the evaluations in Section~\ref{sec:eval}, it can be seen that the proposed setup of decoupling high-resolution image projection from the actual text-conditional generation can be leveraged even more as done in the past \citep{vqgan_esser2021taming,saharia2022photorealistic,dalle2_ramesh2022hierarchical}, while still staying on-par or outperforming in terms of quality, fidelity and alignment. Stage C, being the most expensive stage to train from scratch, required only 24,602 GPU hours, compared to 200,000 GPU hours \citep{sd21-model-card} for \ac{SD} 2.1, making it a 8x improvement.  Additionally, \ac{SD} 1.4 and 2.1 processed significantly more image samples. The latter metric is based on the total number of steps of all trainings and finetunings and multiplied with the respective batch sizes. 
Even when accounting for 11,000 GPU hours and 318M train samples used for training Stage B, Würstchen is significantly more efficient to train than the \ac{SD} models.
Moreover, although needing to sample with both Stage A \& B to generate the VQGAN latents $\mathbf{\bar{x}}_{q}$, the total inference is still significantly faster than \ac{SD} 2.1 and XL (see Figure \ref{fig:inference_speed}).

\section{Conclusion}
In this work, we presented our text-conditional image generation model Würstchen, which employs a three stage process of decoupling text-conditional image generation from high-resolution spaces. The proposed process enables to train large-scale models efficiently, substantially reducing computational requirements, while at the same time providing high-fidelity images. Our trained model achieved comparable performance to models trained using significantly more computational resources, illustrating the viability of this approach and suggesting potential efficient scalability to even larger model parameters. We hope our work can serve as a starting point for further research into a more sustainable and computationally more efficient domain of generative AI and open up more possibilities into training, finetuning \& deploying large-scale models on consumer hardware. We will provide all of our source code, including training-, and inference scripts and trained models on GitHub. 

\ificlrfinal
\section*{Acknowledgements}
The authors wish to express their thanks to Stability AI Inc. for providing generous computational resources for our experiments and LAION gemeinnütziger e.V. for dataset access and support.
This work was supported by a fellowship within the IFI program of the German Academic Exchange Service
(DAAD). 

\section*{Author contributions}
The model architecture was designed by PP and DR. The model training was carried out by PP and DR. The baseline model was trained and implemented by MR. 
The evaluation was carried out by MR and MA. The manuscript was written by PP, DR, MR, CP and MA.

\fi

\section*{Reproducibility Statement}
We release the entire source code of our pipeline, together with the model weights used to generate these results in our GitHub repository. We also include instructions on how to train the model and an inference notebook. As described in Appendix \ref{sect:model_details}, we only used dedublicated publicly available data to train the model. 
The methodology that was used to conduct the study on human preference can be found in Appendix \ref{sec:preference}. The setup of the comparison between other open-source baselines is described in Section \ref{sec:experiments}. We exclusively used open-soruce models with their official repositories and weights, when computing metrics for other models.

\begin{acronym}
\acro{CLIP}[CLIP]{Contrastive Language-Image Pretraining}
\acro{MaskGIT}[MaskGIT]{Masked Generative Image Transformer}
\acro{GAN}[GAN]{Generative Adversarial Network}
\acro{LDM}[LDM]{Latent Diffusion Model}
\acro{VQ-VAE}[VQ-VAE]{Vector-quantized Variational Autoencoder}
\acro{VQGAN}[VQGAN]{Vector-quantized Generative Adversarial Network}
\acro{CFG}[CFG]{Classifier-Free Guidance}
\acro{SD}[SD]{Stable Diffusion}
\acro{FID}[FID]{Fréchet Inception Distance}
\end{acronym}

\bibliography{egbib}
\bibliographystyle{iclr2024_conference}

\newpage
\appendix

\section{Collages (Randomly Chosen from Parti-prompts)}
\label{sec:collages}

\begin{figure}[ht!]
\centering\includegraphics[width=1.0\textwidth]{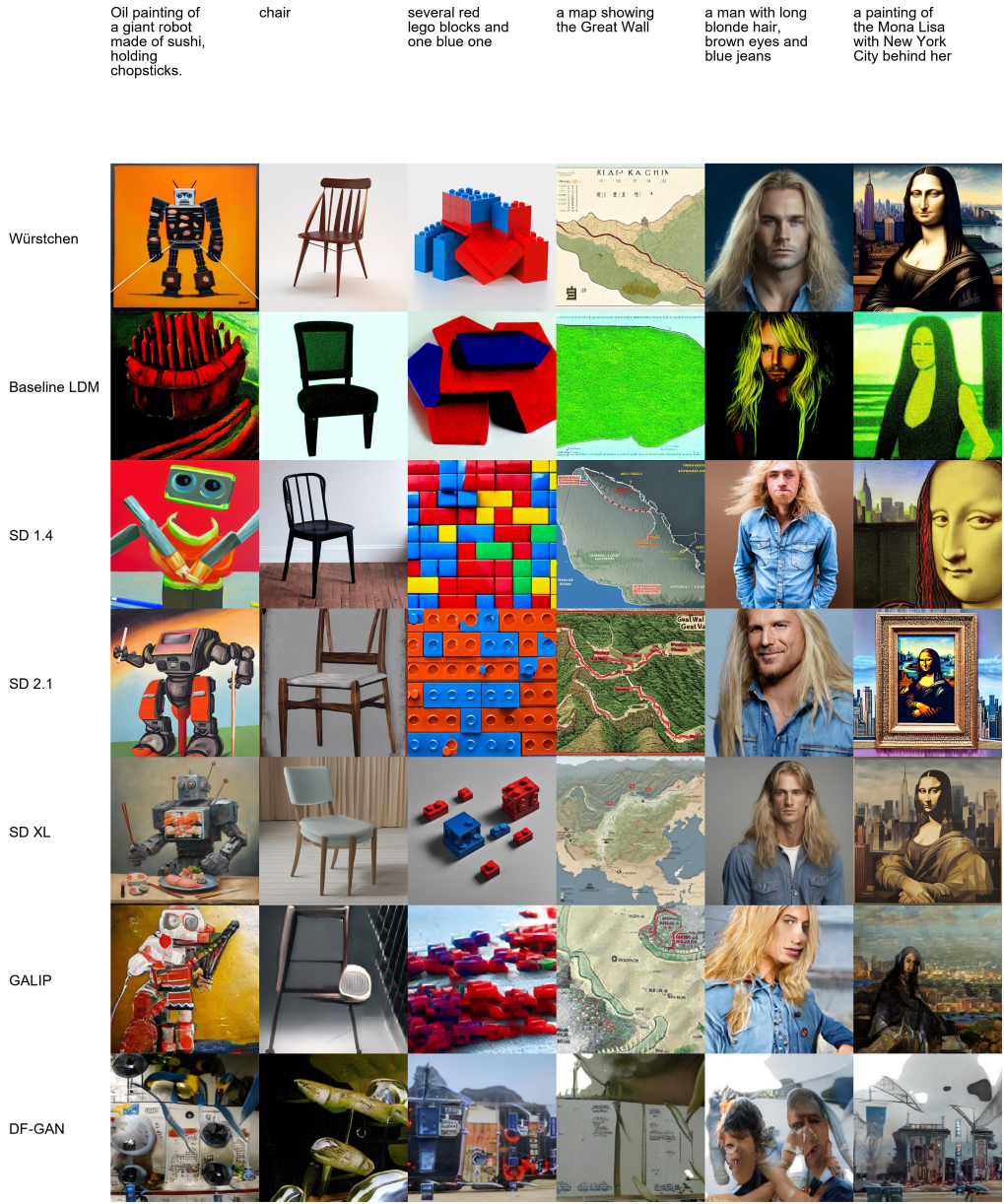}
\caption{Collage \# 1}
\label{fig:collage1}
\end{figure}

\clearpage

\begin{figure}[ht!]
\centering\includegraphics[width=1.0\textwidth]{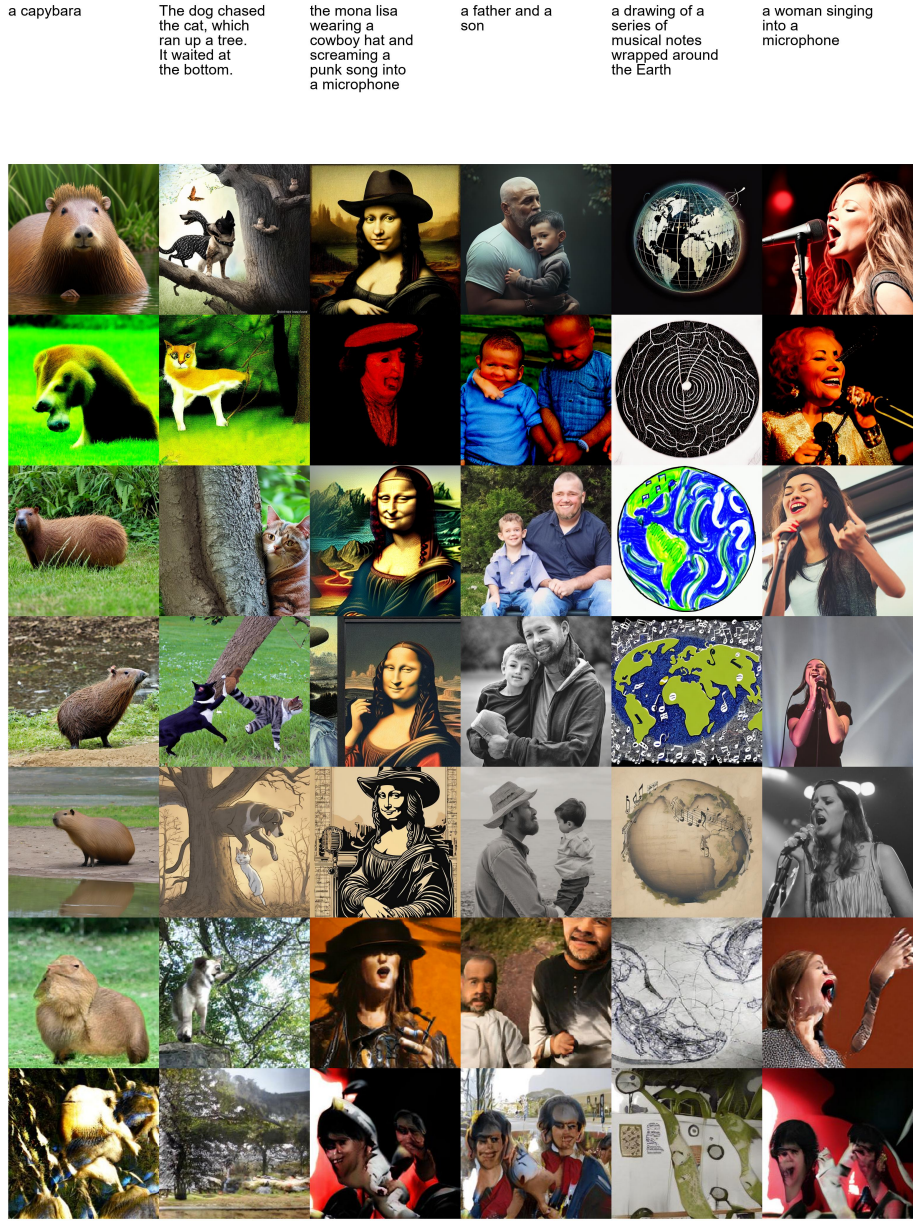}
\caption{Collage \# 2}
\label{fig:collage2}
\end{figure}

\clearpage

\begin{figure}[ht!]
\centering\includegraphics[width=1.0\textwidth]{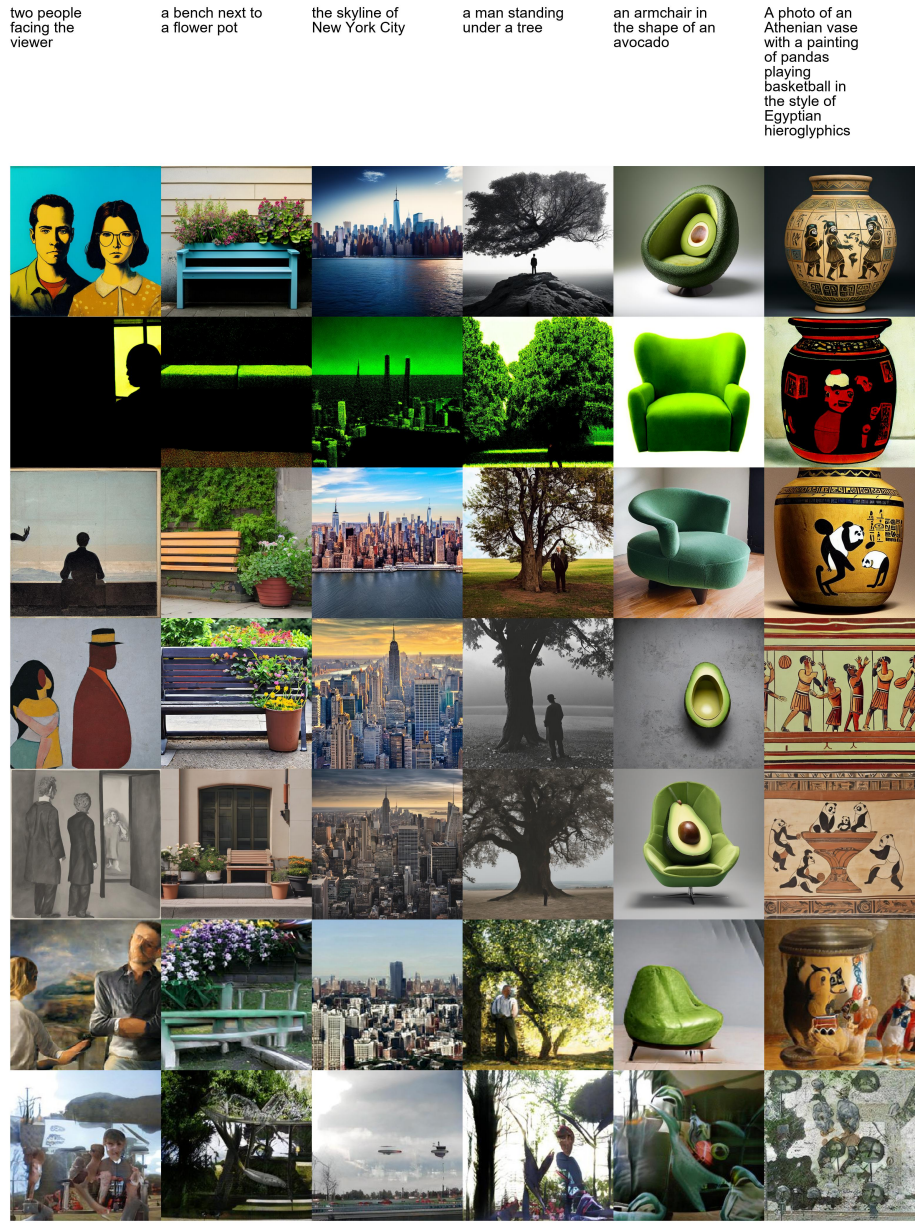}
\caption{Collage \# 3}
\label{fig:collage3}
\end{figure}

\clearpage

\begin{figure}[ht!]
\centering\includegraphics[width=1.0\textwidth]{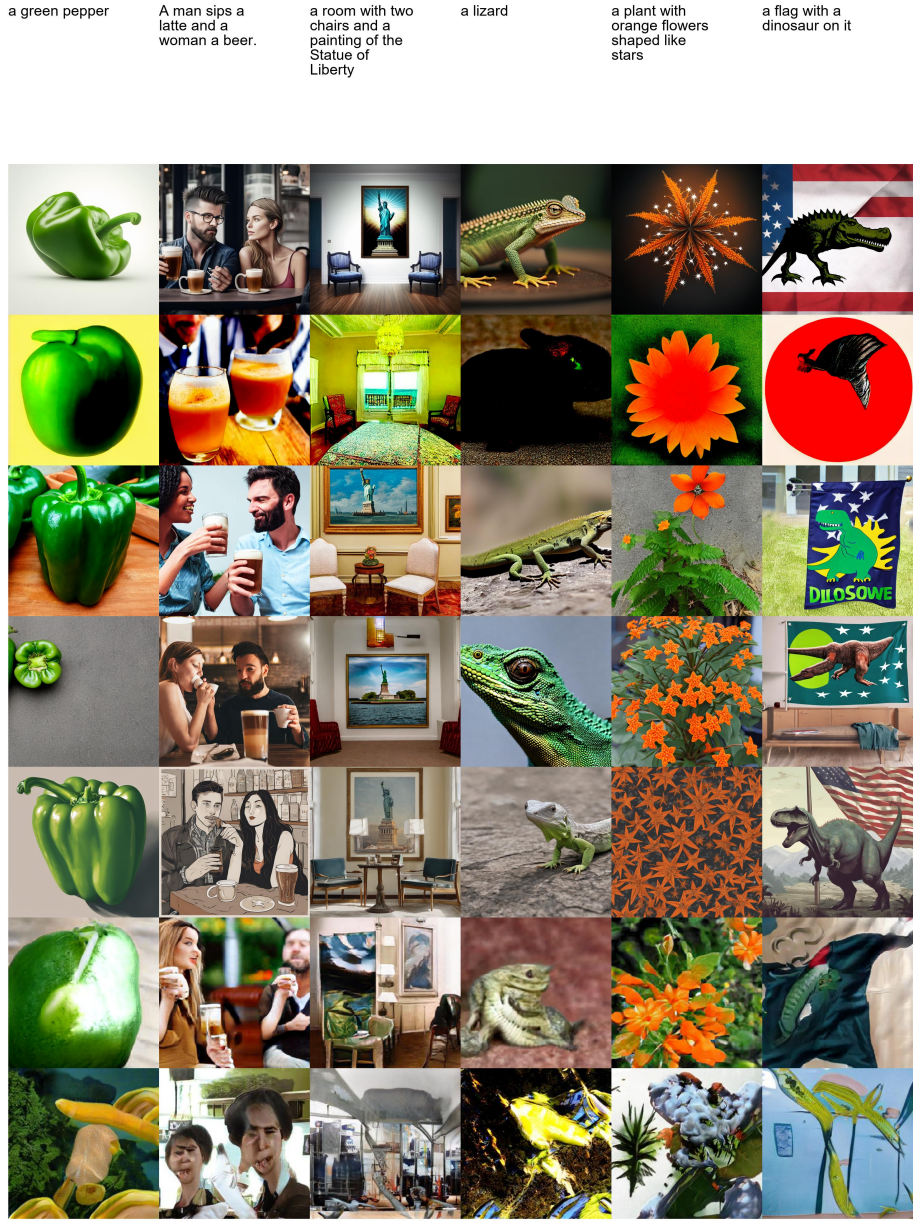}
\caption{Collage \# 4}
\label{fig:collage4}
\end{figure}

\clearpage

\begin{figure}[ht!]
\centering\includegraphics[width=1.0\textwidth]{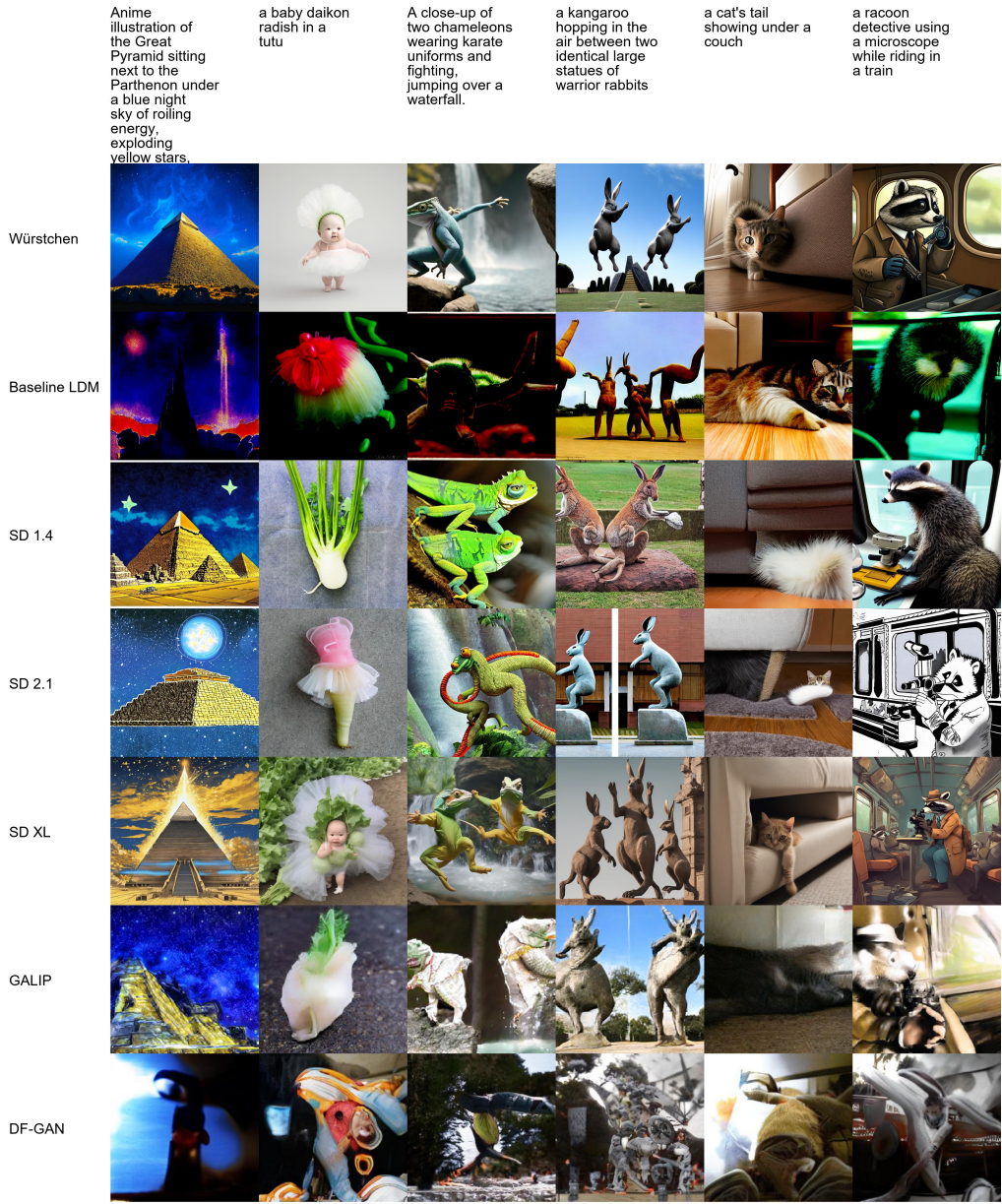}
\caption{Collage \# 5}
\label{fig:collage5}
\end{figure}

\clearpage

\begin{figure}[ht!]
\centering\includegraphics[width=1.0\textwidth]{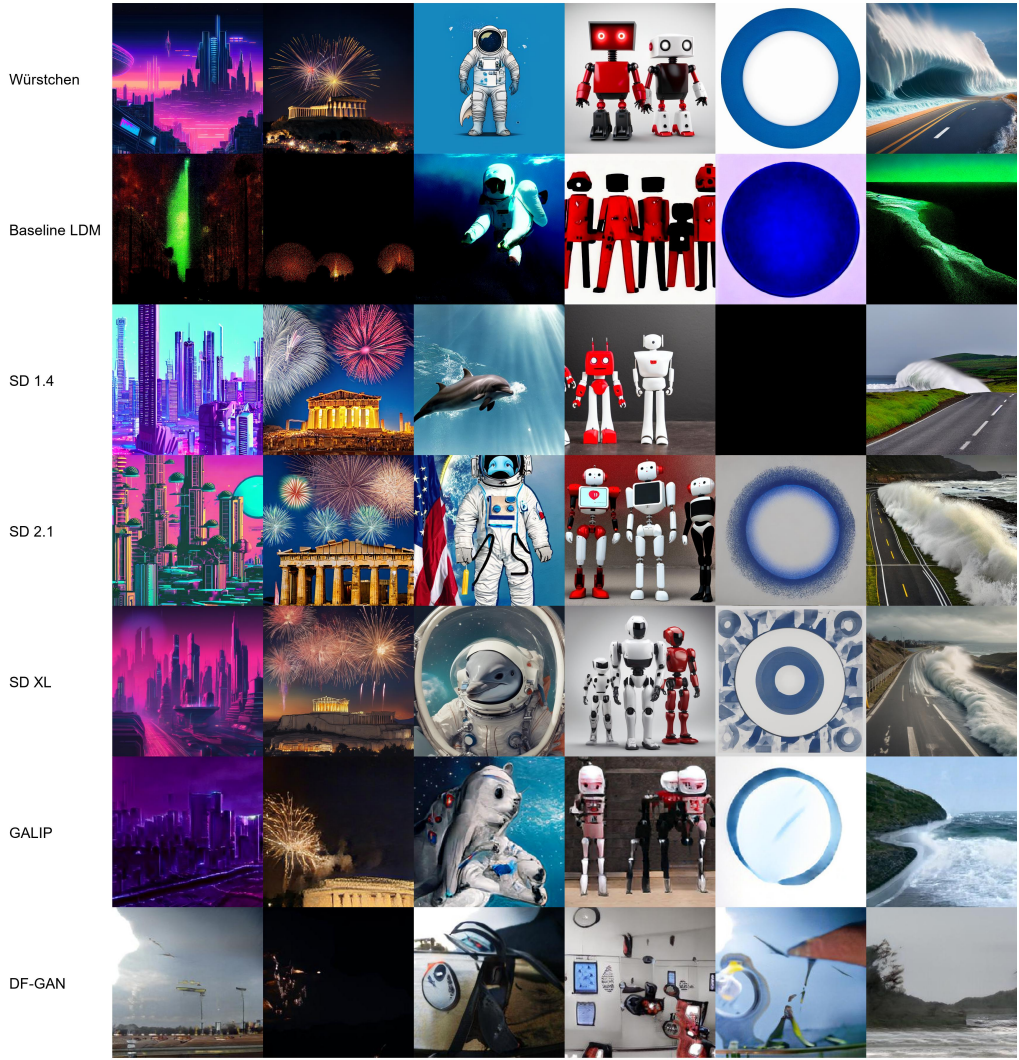}
\caption{Collage \# 6}
\label{fig:collage6}
\end{figure}

\clearpage

\begin{figure}[ht!]
\centering\includegraphics[width=1.0\textwidth]{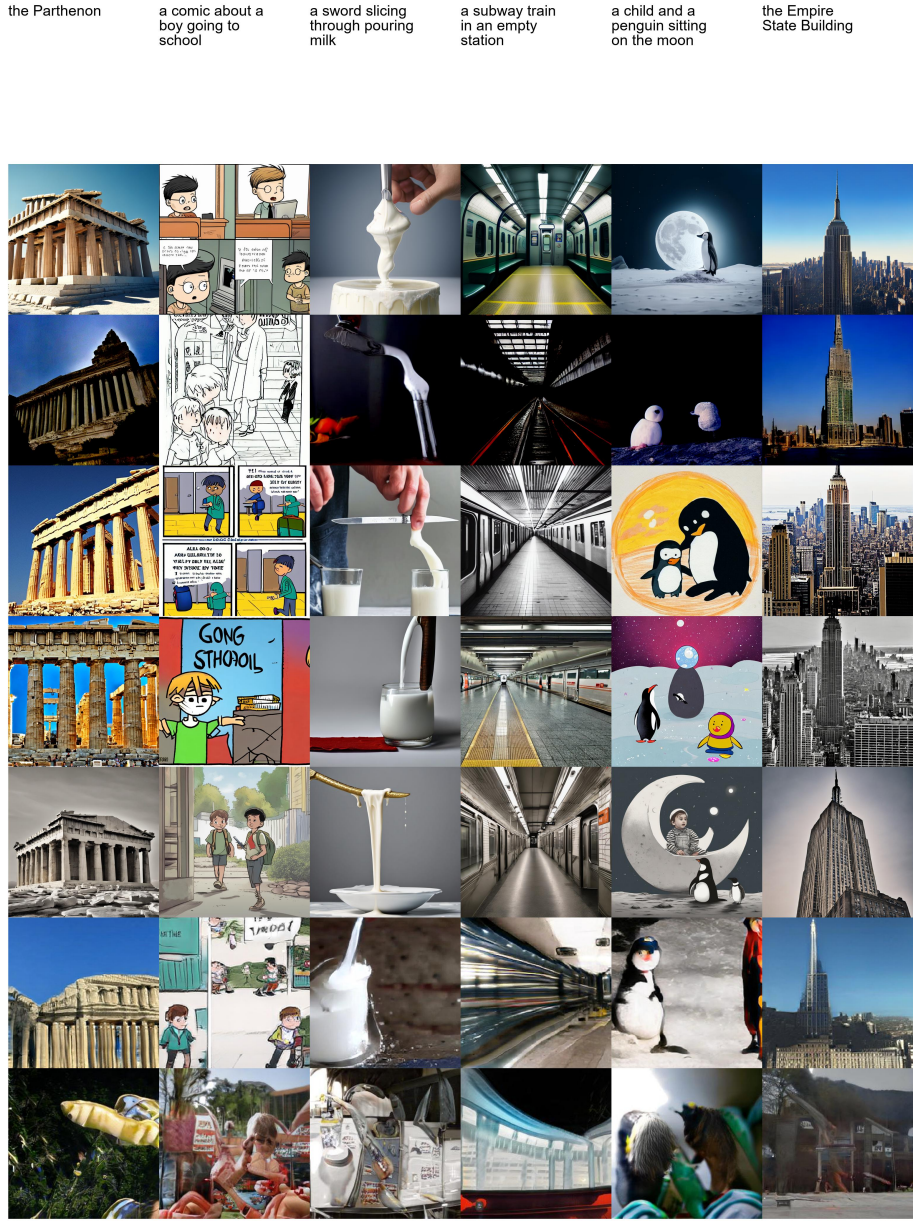}
\caption{Collage \# 7}
\label{fig:collage7}
\end{figure}

\clearpage

\begin{figure}[ht!]
\centering\includegraphics[width=1.0\textwidth]{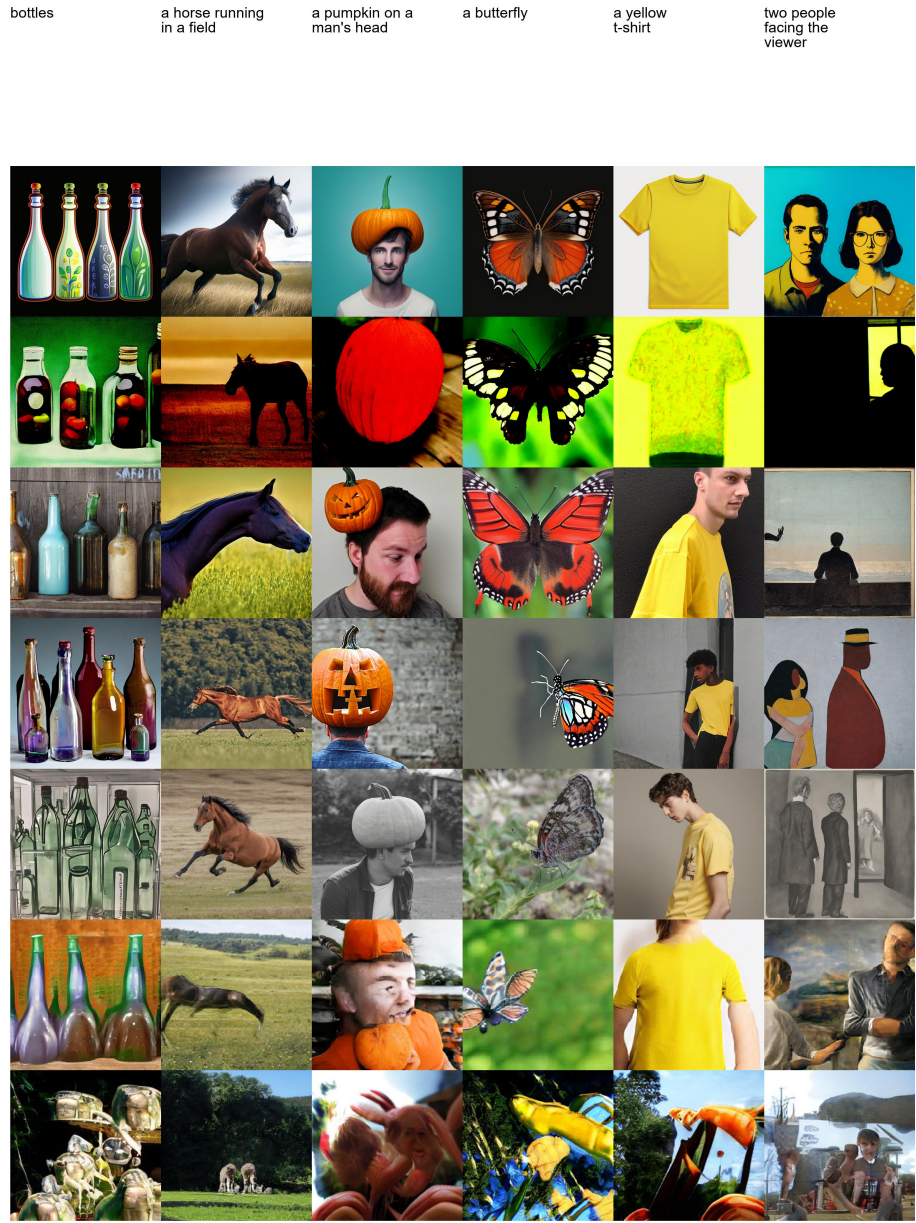}
\caption{Collage \# 8}
\label{fig:collage8}
\end{figure}

\clearpage

\begin{figure}[ht!]
\centering\includegraphics[width=1.0\textwidth]{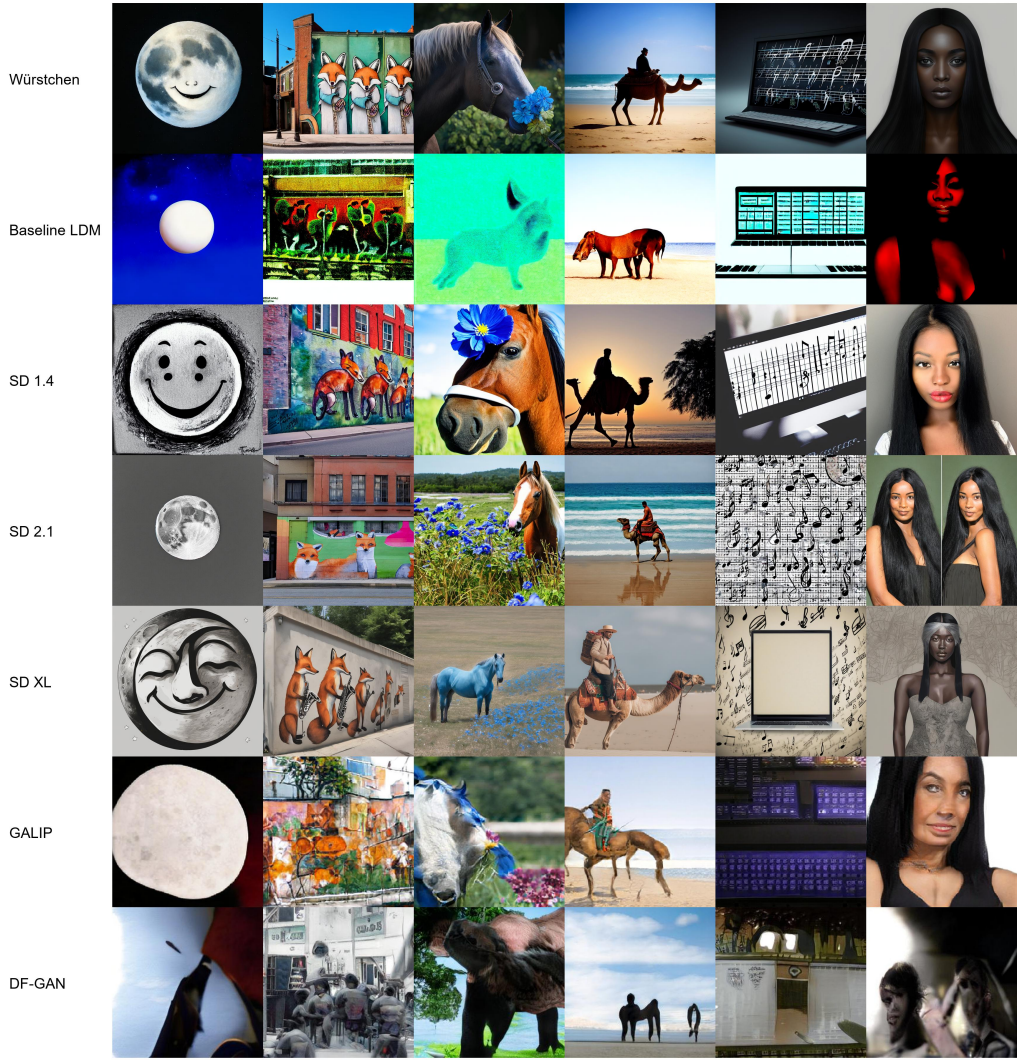}
\caption{Collage \# 9}
\label{fig:collage9}
\end{figure}

\clearpage

\begin{figure}[ht!]
\centering\includegraphics[width=1.0\textwidth]{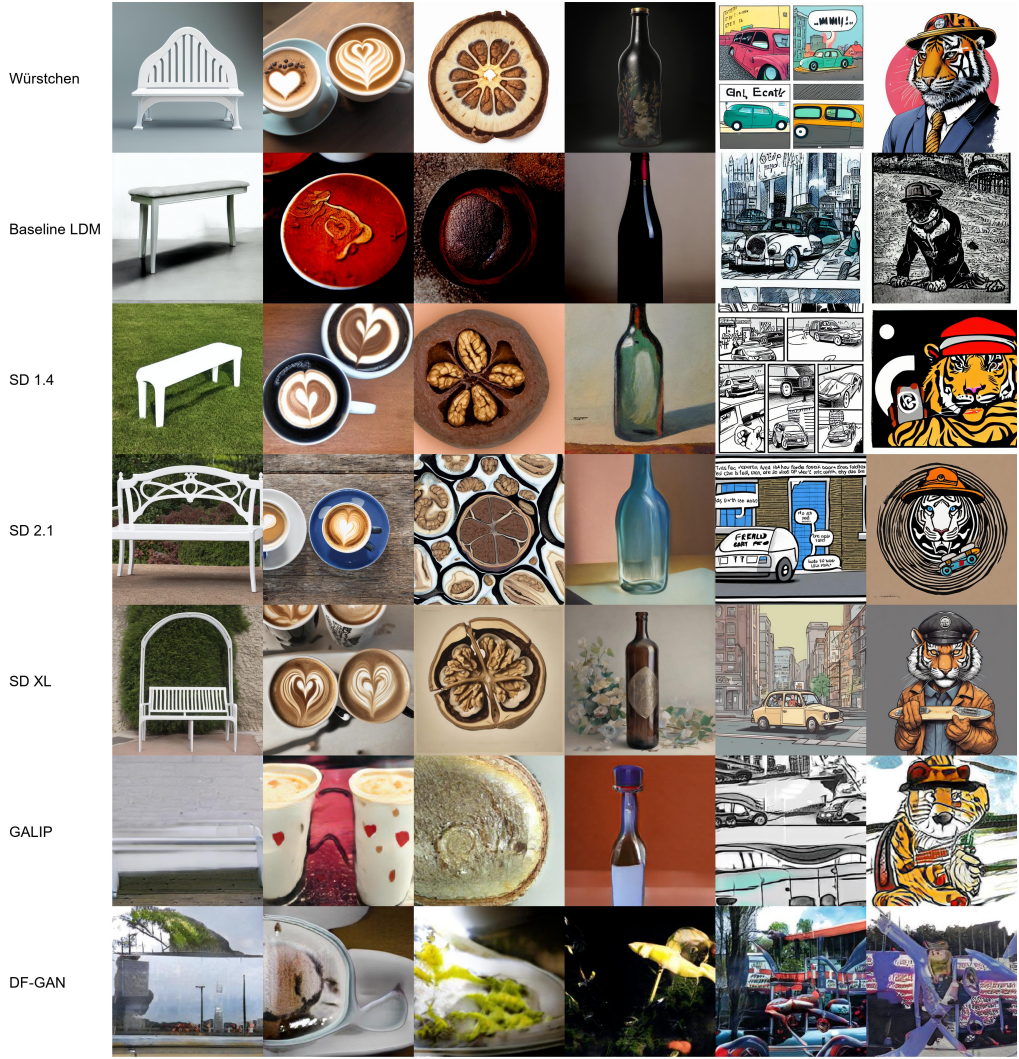}
\caption{Collage \# 10}
\label{fig:collage10}
\end{figure}

\clearpage

\section{Robustness Assessment of Fréchet Inception Distance}
\label{sec:fid}

The \ac{FID} is commonly used to evaluate the fidelity of text-conditional image generators. For this, a large quantity (e.g., 30k) of prompts are retrieved from a dataset that was not used for the training of the system under test. The original images corresponding to those prompts as well as the generated images are then fed to an independent model (typically Inception V3) that was trained on another independent dataset (typically ImageNet). As this model was typically trained on a reduced input resolution (e.g., 299x299 in the case of Inception v3), resampling of the input images is necessary. The FID is calculated from the first and second order statistics of the features found as output of the feature extractor of that model. 

Usage of a metric, however, also implies knowledge of eventual misbehaviors of said metric, hence we compare in this section the reaction of the \ac{FID} towards manipulations that we assumed could be assumed not to be linked towards strong changes in image fidelity. In particular, we stored the images using various quality settings for JPEG compression, used different methods for resampling of the images and performed mild changes in image color, brightness and contrast. 

As shown in Table \ref{tab:fid_robustness}, while changes that are part of standard image augmentation like mild changes in brightness and contrast do not impact the \ac{FID} significantly, a moderate JPEG compression as well as just a change of resampling do impact the metric strongly. In particular, a JPEG compression with 70\% quality, which impedes the image quality only to a small degree, yields an \ac{FID} score in the range of our results, without involving any image generation.   

\begin{table}[h]
\begin{tabular}{lcrr}
Manipulation & Configuration & FID @ COCO 30k & FID @ CelebA-HQ \\
\hline
\multirow{6}{*}{JPEG compression} & quality=95\% & 0.268 & 0.560 \\
& quality=90\% & 1.713 & 2.381 \\
& quality=80\% & 6.658 & 6.291 \\
& quality=70\% & 10.469 & 9.156 \\
& quality=60\% & 13.274 & 11.617 \\
& quality=50\% & 15.129 & 13.519 \\
\hline 
\multirow{2}{*}{Resampling} & NN interpolation & 5.239 & 3.705 \\
& bilinear interpolation & 0.330 & 0.569 \\
\hline
\multirow{5}{*}{Color change} & 8-bit color palette & 27.989  & 31.289 \\
& brightness +10\% & 0.085 & 0.112 \\
& brightness -10\% & 0.054 & 0.101 \\
& contrast +10\% & 0.051 & 0.077 \\
& contrast -10\% & 0.072 & 0.098\\

\end{tabular}
\caption{Assessment of Fréchet Inception Distance (FID) following minor image manipulations that are not expected to significantly alter the fidelity and composition of the images.}
\label{tab:fid_robustness}

\end{table}

\newpage

\clearpage

\section{Methdology of the Human Preference Experiments}
\label{sec:preference}
\paragraph{Used Models:}
The studies were conducted with images generated with \ac{SD} 2.1 and Würstchen.
\paragraph{Data Displayed to User:}
We generated 30,000 images based on the COCO-validation set prompts for each model for the first study and 1,633 images each based on Partiprompts for the second study. 
All images were scanned manually for harmful and graphic and pornographic content.

\paragraph{Setup:}
Both studies are conducted online. Participants are presented an image generated from both models using the same prompt. The prompt is also displayed. Neither the model that generated the images nor the number of models used for the image generation as a whole is known and never displayed to the participants. The displayed images are randomly chosen every time and displayed in a random order.

\begin{figure}[htb!]
\centering\includegraphics[width=\textwidth]{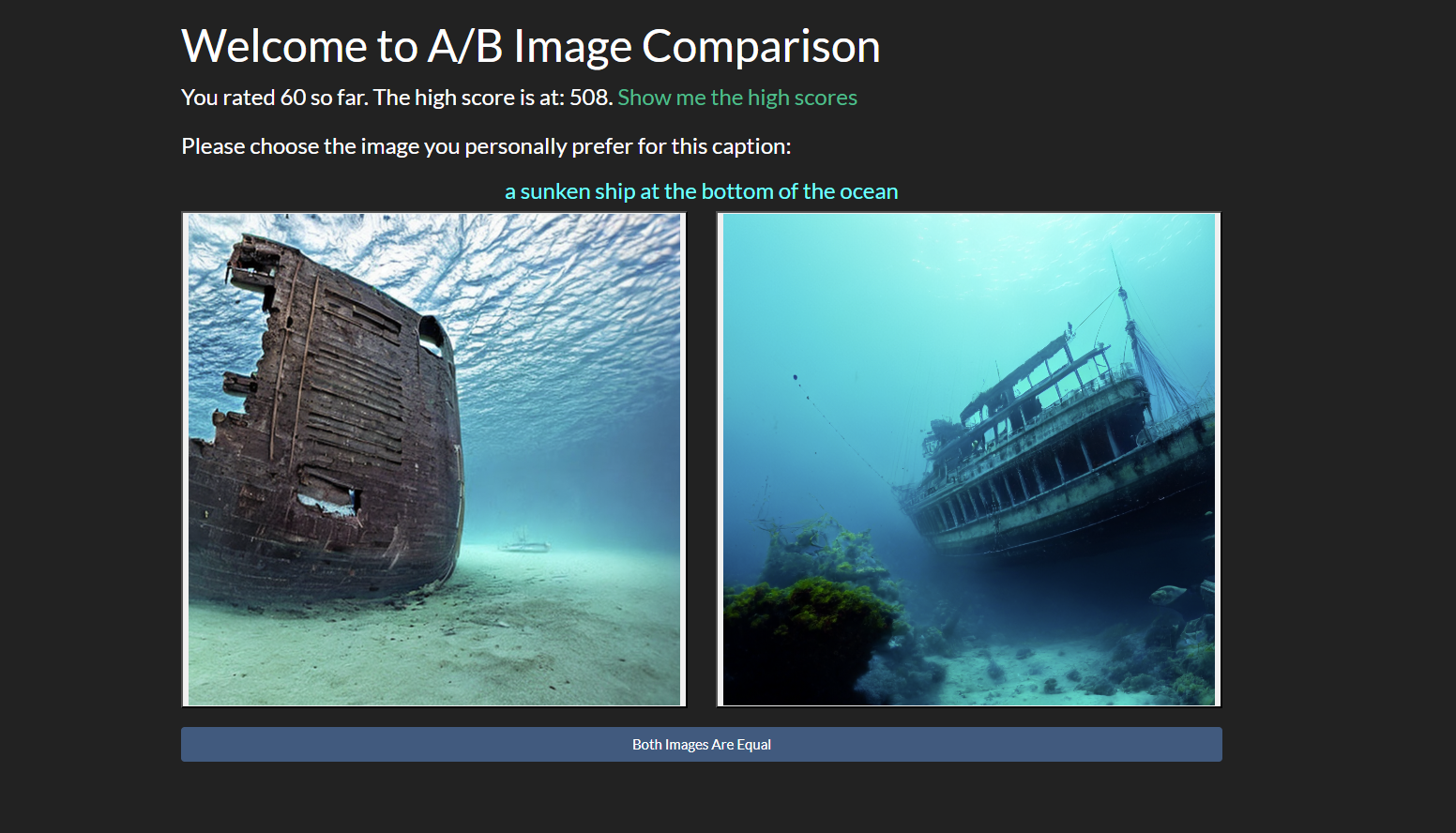}
\caption{A screenshot from the human preference study. Users can click on either image or the button below to select a preference.}
\label{fig:study}
\end{figure}

Participants can answer that they prefer the left, right image or perceive both as equal. 
participants are not paid and are only given the instruction to select images based on the prompt and their personal taste.
The second study, which was conducted on Parti-prompts, also urged the participants to annotate 50 pairs to avoid a long-tailed turnout, which happened in the first study.
For gamification reasons, a leaderboard was added which allows users which achieved a high number of votes to enter a personal alias.

\paragraph{Participation:}
Based on the randomly generated pseudonyms, in total of 90 unique users participated in both studies combined, 33 of which participated exclusively in the first study using images generated from COCO-prompts, 58 participated exclusively in the second study, which used images generated from Parti-prompts, 3 users participated in both studies.

Over both datasets, a total of 4026 comparisons were evaluated. 2490 comparison were done on MS-COCO generated images and 1604 on images generated from Parti-prompt captions.

Participants received no compensation of any form. Participation was pseudonymous.

\clearpage

\section{How are Stage B and C sharing their workload?}
\label{sect:workload}

\begin{figure}
    \centering
    \vspace{-1cm}
    \subfloat[Stage C]{\includegraphics[width=.21\linewidth]{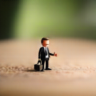}} \hfill
    \subfloat[Stage B and C]
    {\includegraphics[width=.21\linewidth]{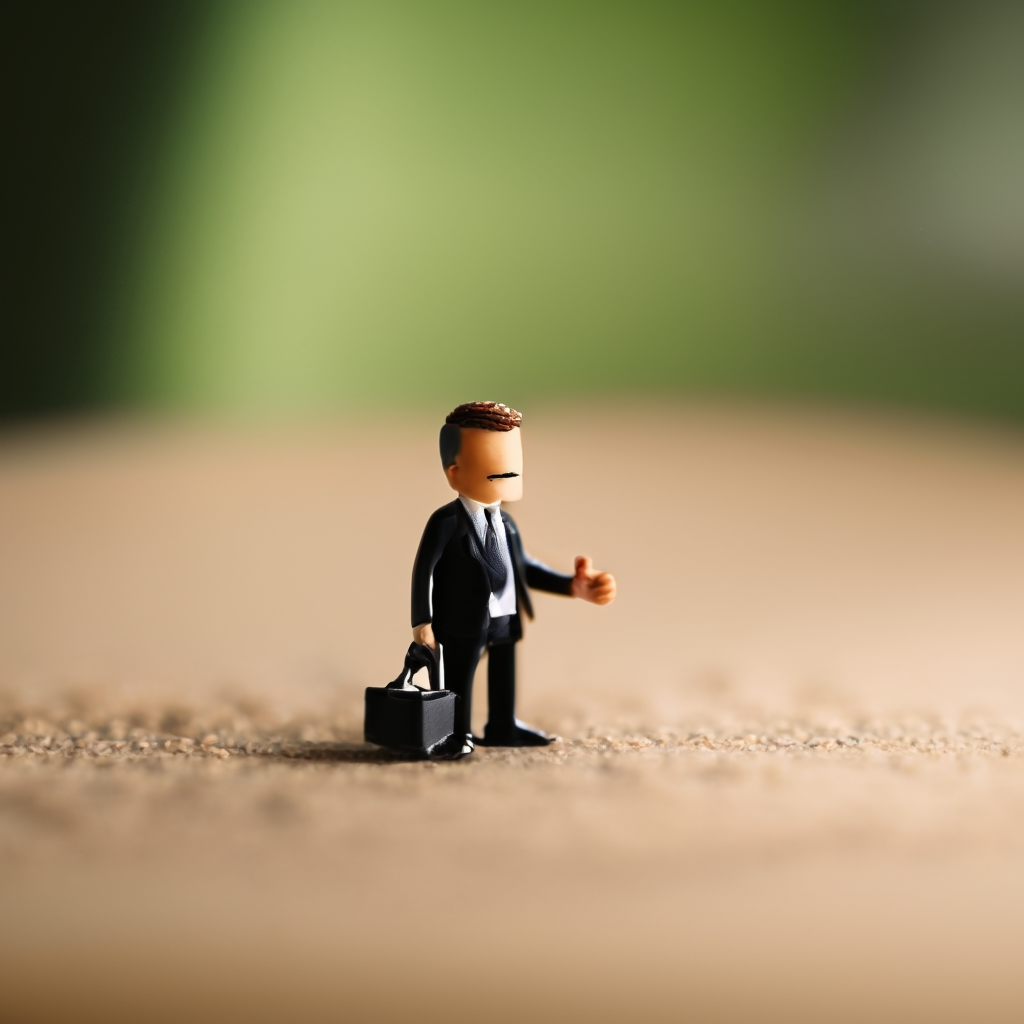}}
    \caption{Caption: Macro photography of a tiny businessman.}

    \vspace{1em} 
    
    \subfloat[Stage C]{\includegraphics[width=.21\linewidth]{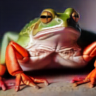}} \hfill
    \subfloat[Stage B and C]{\includegraphics[width=.21\linewidth]{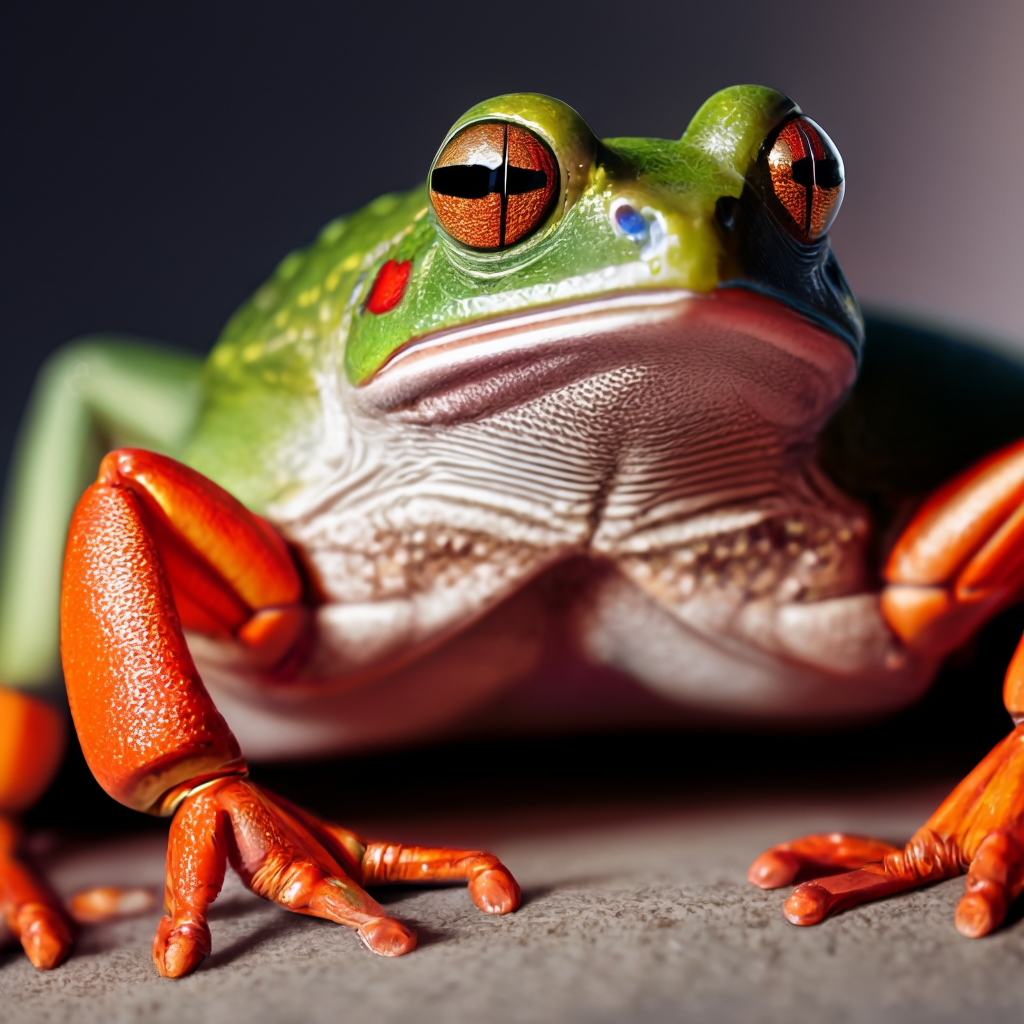}}
    \label{fig:c2}
    \caption{Caption: Dramatic photography of a frog evolving into a crab, crab legs, macro photography.}
    
    \vspace{1em} 
    
    \subfloat[Stage C]{\includegraphics[width=.21\linewidth]{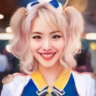}} \hfill
    \subfloat[Stage B and C]{\includegraphics[width=.21\linewidth]{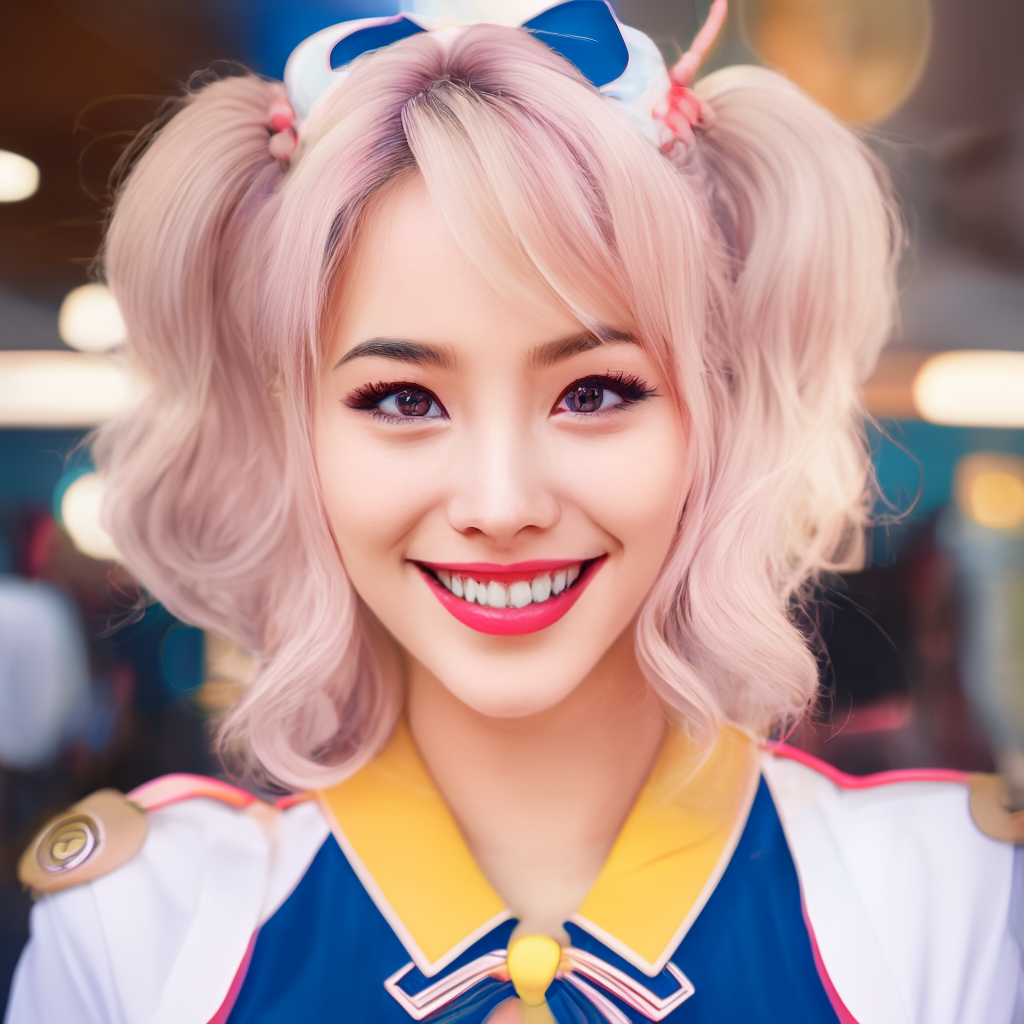}}
    \label{fig:c3}
    \caption{Caption: Cute woman smiling wearing a Sailor Moon cosplay.}
    
    \vspace{1em} 
    
    \subfloat[Stage C]{\includegraphics[width=.21\linewidth]{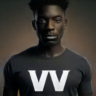}} \hfill
    \subfloat[Stage B and C]{\includegraphics[width=.21\linewidth]{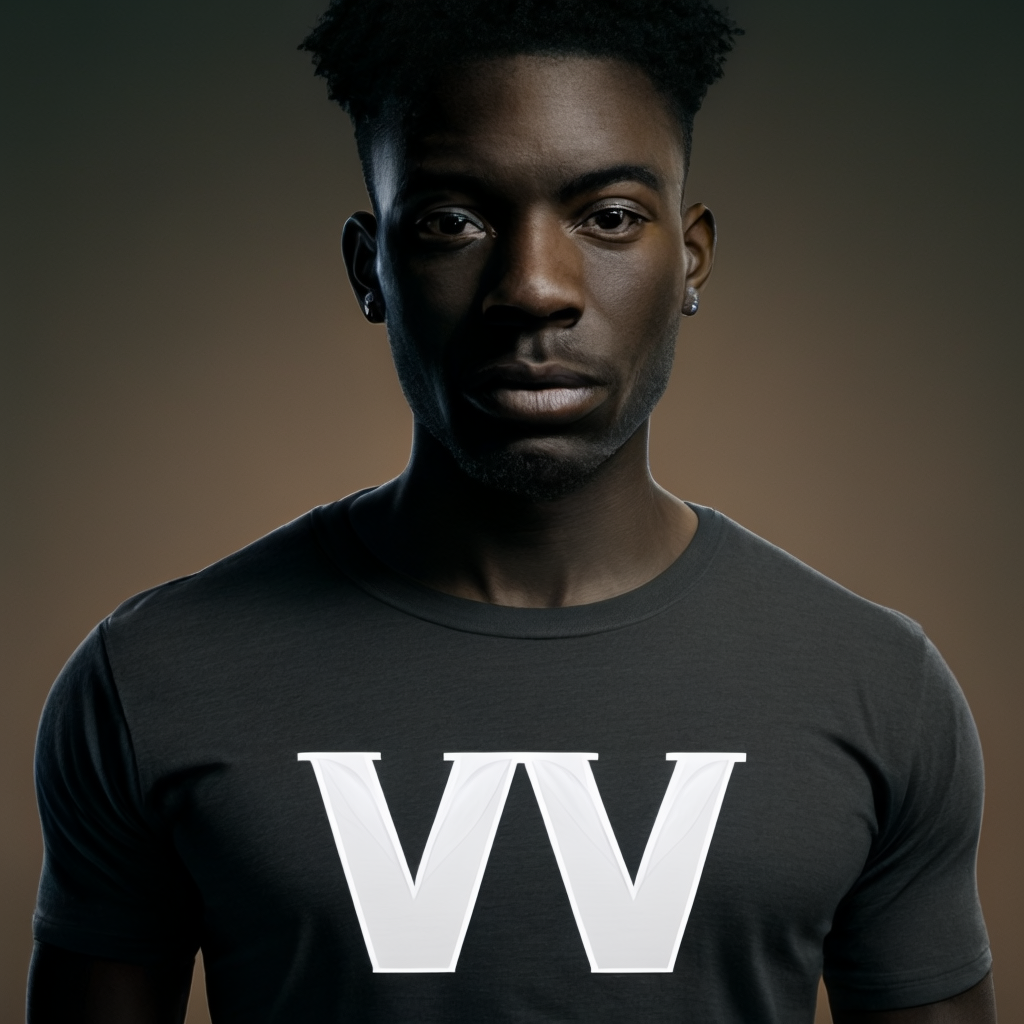}}
    \label{fig:c4}
    \caption{Caption: a black man with a t-shirt with the letter W}

    \subfloat[Stage C]{\includegraphics[width=.21\linewidth]{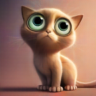}} \hfill
    \subfloat[Stage B and C]{\includegraphics[width=.21\linewidth]{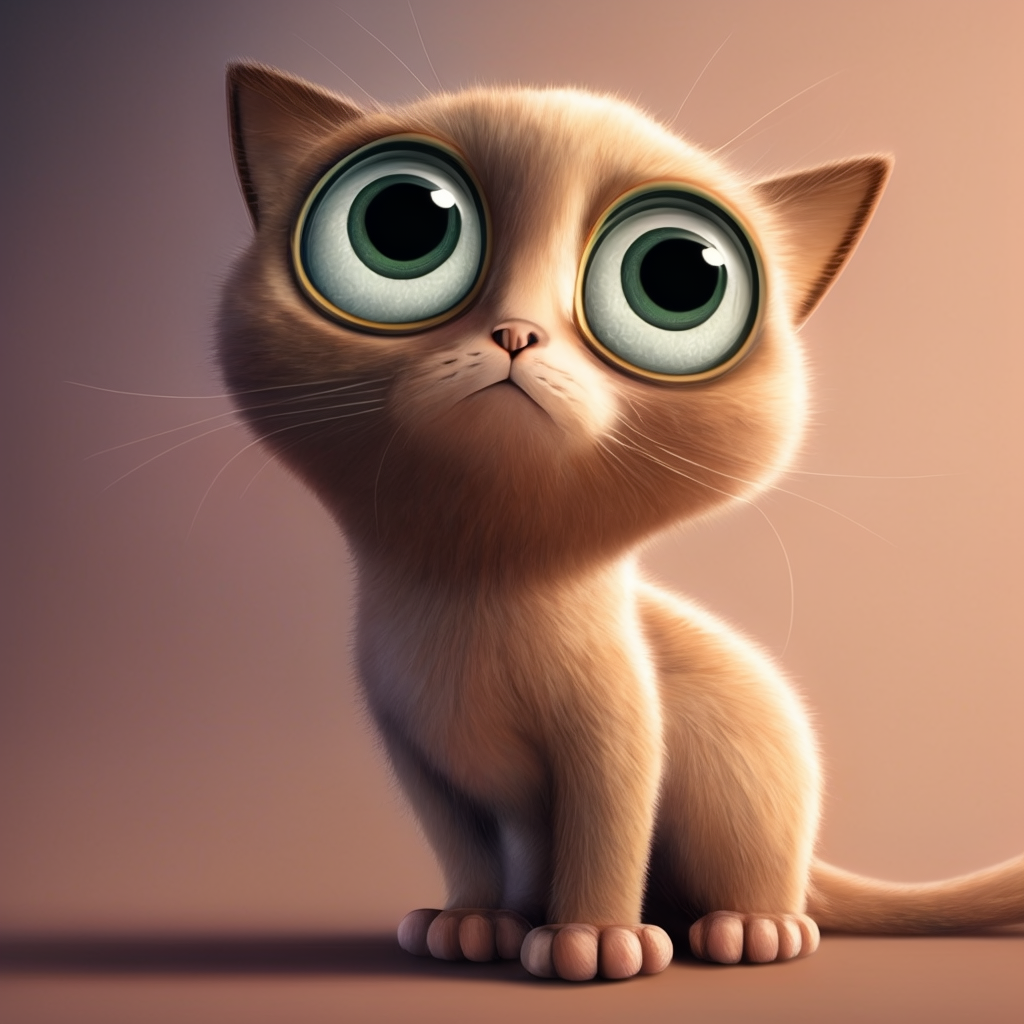}}
    \label{fig:c5}
    \caption{Caption: Cute cat, big eyes pixar style.}
\end{figure}

In our work we view Stage C as the primary working part of the model, when it cames to generating images from text. However, this is not immediately clear from the architecture as Stage B and C are both generative models and have similar capacities.
In this section, we are going to briefly explore how Stage B and Stage C share the workload of image generation.
By doing so, we demonstrate that Stage C is responsible for the content of the image, while Stage B acts as a refining model, adding details and increasing the resolution, but ultimately not changing the image in a semantically meaningful way.
To investigate, we trained a small (3.9M parameter) decoder to reconstruct the images from the latents produced by Stage C and compared the reconstructions with reconstructions from Stage B conditioned on Stage C.
The results in Figures \ref{fig:c2}, \ref{fig:c3}, \ref{fig:c4} and \ref{fig:c5} show that the images generated by Stage C are very similar to the images generated from Stage B and C combined. From the visual inspection we can observe that the main difference are minor details as well as a reduction in blurriness. These changes can be viewed as the main contributions of Stage B.
From this we conclude that Stage C is the primary factor when it comes to transforming text into images.
This is further supported by the fact that short experiments conducted on alternative training regimes suggest that the text conditioning on Stage B does not enhance the quality of the images and could be dropped in future generations our model.

\paragraph{The Decoder Architecture:}

This decoder is very simple, to mitigate the influence on the latents as little as possible, consisting of 4 stages, composed of 2 convolutional layers. the first downsampling layer is $2 \times 2$ convolution with stride size 2.
The second convolution is a $3 \times 3$ convolution with stride size 1 and GELU-activation function and batch norm.
The first stage has 512 channels, consecutive stages half the channel width.
A final $1 \times 1$ convolution squeezes the channels to 3 color channels.

\clearpage

\section{A Detailed Look at the Neural Architectures}
\label{sect:model_details}
\subsection{Stage A}

\paragraph{Neural Architecture:} The VQGAN is composed of an encoder and decoder with 2 stages each, separated with a downsampling or upsampling layer with a $4 \times 4$-kernel and stride size 2.
The encoder starts, and the decoder ends with a pixel shuffle operation with a scale factor of 2.
In the encoder, each stage consists of a single ConvNeXt block \citep{liu2022convnet} with 384 input channels and 1536 embedding channels.
the final layer of the encoders is a $1 \times 1$ convolution followed by a BatchNorm-layer, reducing the number of channels to the dimensionality of the encoding layer (4).
The decoder reverses this operation with a similar combination of layers.
The decoding layer uses 16 blocks in the first stage and 1 block in the second stage. All blocks have 384 input channels and 1536 embedding channels.

\paragraph{Training Details:}
The model is trained in a single run with 500,000 iterations using the AdamW optimizer with a learning rate of $1e-4$ and a batch size of 256.
The model is fed with $128\times128$ pixel crops from images taken from the deduplicated subsets of the improved-aesthetic LAION-5B \citep{schuhmann2022laion} dataset, which were previously resized to $256 \times 256$ images.
As we anticipate the removal of the quantization, we randomly drop the quantization during training with a drop-chance 10\%.
For training we use three distinct losses. Mean Squared Error (MSE), Adverserial Loss (AL) and Perceptual Loss (PL).
for the first 10.000 iterations we use the loss weights of (1.0, 0.0, 0.1) ofr MSE, AL and PL respectively. We activate the AL after 10,000 training steps by increasing the weight to 0.01.

\subsection{Stage B}

\paragraph{Neural Architecture:} Stage B is a U-Net architecture with 4 stages in the encoder and decoder on the latents of the (unquantized) latent of the Stage A VQGAN.
The stages have a channel width of 320, 640, 1280 and 1280 respectively. Each stage starts with a single convolutional layer with a $2 \times 2$ kernel and a stride size of 2 acting as downsampling layer. 
The stages consist of a number of building blocks. Except for the first stage, a building block consists of a ConvNeXt block \citep{liu2022convnet}, a time-step block, applying a linear conditioning \citep{rombach2022high} on the latents and a cross-attention block for conditioning on text-embeddings and image-embeddings. The first stage omits the cross-attention for conditioning on text and images. GlobalResponseNorm and GELU activation functions are used for normalization and as activation functions respectively. The cross-attention mechanisms of each stage have a different number of attention-heads: -, 10, 20, 20 (as the first stage is only conditioned on time).
For all stages with cross-attention, the latents of Semantic Compressor are also concatenated to each residual block before the channelwise-convolution is applied. To fit the respective feature-map size, bicubic interpolation is used to resize the latent of the Semantic Compressor.
The encoder and decoder stages have 4, 4, 14 and 4 blocks respectively. The clip-embeddings have a dimensionality of 1024. The Semantic Compressor produces latents with 16 channels.

The Semantic Compressor is composed of an EfficientNetV2 S \citep{tan2021efficientnetv2} backbone. The final global-pooling and classification-head is replaced by a $1 \times 1$-convolution with stride size $1$, compressing the channels down to 16. It is worth noting that EfficientNetV2 S is dropped after the training of Stage B and C is complete, as it is replaced by Stage C during inference time.

\paragraph{Training Details:}
The model is trained using the AdamW optimizer \citep{https://doi.org/10.48550/arxiv.1711.05101} with a learning rate of $1e^{-4}$ using a linear warm-up schedule for 10k steps.
In total, the model is trained for 457,000 iterations with an input resolution of $512 \times 512$ and a batch size of 512. The model is trained for an additional 300,000 iterations with an input resolution of $1024 \times 1024$ and a batch size of 128. The resolutions described in this work are the image sizes fed into the VQGAN for encoding. The Semantic Compressor is fed an input resolution of $384 \times 384$ pixels and  $768 \times 768$ during these two phases of training respectively. The training time and compute budget listed in the paper reflects this entire training process. 
All images fed into the semantic compressor are normalized using $\mu=(0.485, 0.456, 0.406)$ and $\sigma=(0.229, 0.224, 0.225)$.
The model is trained on a deduplicated subsets of the improved-aesthetic LAION-5B \citep{schuhmann2022laion} dataset.

\subsection{Stage C}

\paragraph{Neural Architecture:}
Stage C consists of a sequence of 16 building blocks. Each building block is composed of ConvNeXt block \citep{liu2022convnet} a time-conditioning block and a cross-attention block for text-conditioning, similar to Stage B. Text-Conditioning is applied from an unpooled CLIP-H model.
Text-embedding has a dimensionality of 1024, each cross-attention block has 16 heads. The width of the network is 1280 channels.

The Semantic Compressor is composed of an EfficientNetV2 S \citep{tan2021efficientnetv2} backbone trained during Stage B training and otherwise unchanged. It is worth noting that EfficientNetV2 S is dropped after the training of Stage B and C is complete. During inference time, the output of Stage C instead used to condition Stage B, replacing the Semantic Compressor entirely. The Semantic Compressor is not trained during Stage B training.

\paragraph{Training Details:}
The model is trained a total of 4 consecutive times using the AdamW optimizer with a learning rate of $1e-4$. The first three consecutive trainings are conducted on a deduplicated subsets of the improved-aesthetic LAION-5B \citep{schuhmann2022laion} dataset. The final training is conducted on the dataset but further filtered by aesthetical artworks.
The images are fed into the frozen Semantic Compressor to produce latents of a specific resolution.
We provide the resolution of the latents alongside the resolution fed into the Semantic Compressor. The preprocessing done on images of the compressor is identical to Stage B.

The first training is conducted for 500,000 iterations on $12 \times 12$ latents, which corresponds to an $384 \times 384$ input resolution for the semantic compressor using a batch size of 1536.
The second training is conducted for an additional 364,000 iterations on $24 \times 24$ latents, corresponding to $768 \times 768$ images being fed into the Semantic Compressor, using a batch size of 1536.
The third training is run for only 4,000 steps and is done to adapt the model to various aspect ratios. The aspect ratio is randomized uniformly for each batch of 768 images to one of the three following values: $768 \times 1280$, $1280 \times 768$ and $768 \times 768$.
The fourth and final training is designed to improve the aesthetical quality of images and is conducted for another 50,000 iterations using a batch size of 384 and a resolution of $768 \times 768$ ($24 \times 24$ latents).

The final model is a 50:50 interpolation between the weights after the 3rd training and the final training run. This allows the model to generate a blend of aesthetic/artistic and realistic images. However, we open source the two models this interpolation is based on besides this final model.

\subsection{Baseline LDM}

\paragraph{Neural Architecture:} The \ac{LDM} is a U-Net architecture with 4 stages in the encoder and decoder on the latents of the VAE used by Stable-Diffusion 1.4.
The stages have a channel width of 320, 640, 1280 and 1280 respectively. Each stage starts with a single convolutional layer with a $2 \times 2$ kernel and a stride size of 2 acting as downsampling layer. 
The stages consist of a number of building blocks. A building block consists of a ConvNeXt block \citep{resnet_he2016deep} and two cross attention blocks for conditioning on time and text-embeddings, using GlobalResponseNorm and GELU activation functions. The first cross-attention in a building block conditions on the time step $t$, while the second one on the text embeddings $c_{text}$. The cross-attention mechanisms of each stage have a different number of attention-heads: 5, 10, 20, 20.
The encoder stages have 2, 4, 14 and 4 blocks respectively, while the corresponding decoder stages have 5, 15, 5 and 3 blocks. Like for the other models, the clip-embeddings have a dimensionality of 768.
Dropout is applied with a probability of 10\% on a features of the text and image embeddings as well as the $3\times 3$-convolution in the ConvNeXt-block.

\paragraph{Training Setup:} The model is trained using the AdamW optimizer \citep{https://doi.org/10.48550/arxiv.1711.05101} with a learning rate of $1e^{-4}$ using a linear warm-up schedule for 10k steps and a batch size of 1280. The training is conducted for approximatly 25,000 GPU hours, which roughly corresponds to 1.5 million training steps.

The model is trained on subsets of the improved-aesthetic LAION-5B \citep{schuhmann2022laion} dataset.
We use a dropout of 5\% on the CLIP-H-text embeddings.

\end{document}